\documentclass[runningheads]{llncs}
\usepackage{graphicx}

\usepackage{tikz}
\usepackage{comment}
\usepackage{amsmath,amssymb} 
\usepackage{color}

\usepackage[accsupp]{axessibility}  

\usepackage[capitalize]{cleveref}
\crefname{section}{Sec.}{Secs.}
\Crefname{section}{Section}{Sections}
\Crefname{table}{Table}{Tables}
\crefname{table}{Tab.}{Tabs.}

\usepackage{booktabs}
\usepackage{multirow}
\usepackage{subcaption}

\begin{document}
\pagestyle{headings}
\mainmatter
\def\ECCVSubNumber{2337}  

\title{Symmetry Regularization and Saturating Nonlinearity for Robust Quantization} 

\titlerunning{SymReg and SatNL for Robust Quantization}
%

\author{Sein Park\inst{1}* \and
Yeongsang Jang\inst{2}* \and
Eunhyeok Park\inst{1,2} \\
\text{\footnotesize* equal contribution}
}%
\authorrunning{S. Park* et al.}
%
\institute{
Graduate School of Artificial Intelligence,\and
Department of Computer Science and Engineering \newline
POSTECH, Pohang, Korea \newline
\email{\{seinpark, jangys, eh.park\}@postech.ac.kr}
}


\maketitle

\begin{abstract}
Robust quantization improves the tolerance of networks for various implementations, allowing reliable output in different bit-widths or fragmented low-precision arithmetic. 
In this work, we perform extensive analyses to identify the sources of quantization error and present three insights to robustify a network against quantization: reduction of error propagation, range clamping for error minimization, and inherited robustness against quantization. Based on these insights, we propose two novel methods called symmetry regularization (SymReg) and saturating nonlinearity (SatNL). Applying the proposed methods during training can enhance the robustness of arbitrary neural networks against quantization on existing post-training quantization (PTQ) and quantization-aware training (QAT) algorithms and enables us to obtain a single weight flexible enough to maintain the output quality under various conditions. We conduct extensive studies on CIFAR and ImageNet datasets 
and validate the effectiveness of the proposed methods.
\keywords{Robust Quantization, Post-training Quantization (PTQ), Quantization-aware Training (QAT)}
\end{abstract}

\section{Introduction}

Deep learning algorithms have shown excellence in diverse applications, but the increasing memory footprint and computation overhead have become obstacles to utilizing them. To exploit the excellence of deep neural networks (DNNs) in practice, neural network optimization is becoming more and more important. 
Neural network quantization is a representative optimization technique beneficial to footprint reduction and performance improvement. Due to its practical advantages, advanced hardware is already equipped with low-precision support, such as the well-known float16, bfloat16, and int8-based operations~\cite{NVIDIA_8bit,NVIDIA_bfloat,TPUv4,IntArithmetic,SNAP,song20197}, even with 4-bit or lower-precision acceleration~\cite{OLAccel,BitFusion,Andrew,NVIDIA_4bit,Samsung_NPU}. With the aid of a judiciously designed quantization algorithm, we could enjoy the benefit of low-precision computation in reality.

However, quantization has the substantial limitation of accuracy degradation due to the limited representation capability.
Many studies have been actively proposed to address this problem, and Quantization-aware training (QAT) is a representative approach where end-to-end training is applied to refine the simulated error of pseudo (or fake-) quantization~\cite{Esser2020lsq,choi2018pact,park2020profit,Jung2019LearningTQ,Lee2018QuantizationFR,ZhoWu16Dorefa}.
QAT is advantageous in the sense of minimal accuracy degradation in the given bit-width. Recently, post-training quantization (PTQ) has emerged as an alternative approach that quantizes the pre-trained network without fine-tuning~\cite{banner2018aciq,Banner2019PostT4,Zhao2019ImprovingNN,finkelstein2019fighting,Lee2018QuantizationFR,Nahshan2021LossAP,nagel2019data,Cai2020ZeroQAN,AdaQuant,qdrop}.
PTQ allows us to exploit the benefit of low-precision computation with a minimal number of training datasets, thereby having many more practical use cases compared to QAT.

Nonetheless, both QAT and PTQ have severe drawbacks: QAT requires access to the entire dataset and the expenses of an additional training stage. In addition, the model with QAT is specialized for the target precision and quantization scheme, thereby lacking robustness in different bit-widths. 
On the other hand, PTQ suffers from notable accuracy degradation to QAT due to the lower degree of freedom and insufficient information to compensate for the errors.
Many studies have been proposed to overcome this limitation, but a bit-width of 8- or more is still required for the advanced networks~\cite{Cai2020ZeroQAN,AdaQuant,qdrop}.

Recently, alternative approaches have been proposed to enhance the robustness of networks against quantization~\cite{yu2020low,han2021improving,hoffman2019robust,shkolnik2020robust}. 
Improving the robustness of networks has diverse advantages, including allowing the quantized model to maintain the accuracy in bit-widths other than in which it trained and preserving the quality of output with various quantization algorithms.
Practically, these properties help to utilize the pareto-front optimal points of energy consumption and computation, such as exploiting a high-precision model when the resource (e.g., battery) is sufficient and reducing precision dynamically when the resource is scarce. In addition, numerous companies are now designing their own accelerators having divergent and fragmented low-precision implementations. When we need to support multiple accelerators, preparing a single low-precision model robust enough to endure the minor modification of different implementations could be an attractive option for rapid deployment. Robust quantization enables diverse appealing applications, having strong importance in practice. 


In this work, we propose two novel methods to increase the robustness of neural networks based on three insights about the error component of quantization. The paper is organized as follows; first, we perform an extensive analyses to identify the source of errors from quantization and indicate three motivations to robustify the network: reduction of error propagation, range clamping for error minimization, and inherited robustness against quantization (\cref{sec:motiv}). To address those motivations, we introduce two novel ideas: symmetry regularization (SymReg) for the reduction of error propagation and saturating nonlinearity (SatNL) for the others (\cref{sec:imple}). According to our extensive experiments, the proposed methods are beneficial for maximizing the robustness of networks after QAT or PTQ, showing state-of-the-art results. (\cref{sec:exp}). We then clarify the limitation of this study in \cref{sec:limit} and conclude the paper in \cref{sec:con}.

\section{Related Work}
\subsection{Quantization-aware Training and \\Post-training Quantization}
QAT~\cite{Esser2020lsq,choi2018pact,park2020profit,Jung2019LearningTQ,Lee2018QuantizationFR,ZhoWu16Dorefa} shows the potential of low-precision computation, where the milestone networks (e.g., VGG\cite{Simonyan15}, GoogleNet\cite{43022}, and ResNet\cite{ResNet}) could be quantized into sub-4-bit without accuracy loss~\cite{Esser2020lsq,choi2018pact}, and advanced light-weight networks (e.g., MobileNet-V2) could be quantized into 4-bit with negligible accuracy loss~\cite{park2020profit}. Meanwhile, PTQ~\cite{banner2018aciq,Banner2019PostT4,Zhao2019ImprovingNN,finkelstein2019fighting,Lee2018QuantizationFR,Nahshan2021LossAP,nagel2019data,Cai2020ZeroQAN,AdaQuant} applies a conservative bit-width to maintain the quality of output, even though it offers performance benefits with relaxed constraints. In this work, we aim to maximize the benefits of QAT and PTQ through the advantages of robust quantization. 

\subsection{Robustness of Neural Networks}
A line of work closely related to ours is the analysis of the robustness of neural networks, which has attracted attention recently. Currently, the Hessian-aware metric is often used to identify the robustness of neural networks. Previous studies  pointed out that the second derivative of loss is a good approximation of network sensitivity~\cite{dong2019hawq,alizadeh2020gradient}, proposed a way to estimate the approximate Hessian metric efficiently, and showed the potential of sensitivity-aware quantization~\cite{dong2019hawq,dong2019hawqv2,yao2021hawqv3,li2021brecq,qdrop}. On the other hand, few studies have focused on easing the sensitivity of networks during the training phase to improve their generalization performance~\cite{foret2020sam,kwon2021asam}.
The proposed (adaptive) sharpness-aware minimization, (A)SAM, makes the network have lower Hessian spectra than the networks trained without it.
It is expected to be beneficial for enhancing the endurance of the network for quantization, but we observe that the benefit of (A)SAM is degraded in the quantization domain. In this paper, we propose a novel idea, SatNL, to maximize the robustness of quantization with (A)SAM training. 

Moreover, few studies have tried to minimize quantization error in the view of robustness.  GDRQ~\cite{yu2020low} and BR~\cite{han2021improving} attempted to improve the robustness of networks via regularization in QAT tasks. Gradient $l_1$-regularization~\cite{hoffman2019robust} lowered the sensitivity of networks for quantization via regularizing $l_1$-norm of gradient, and KURE~\cite{shkolnik2020robust} regularized the weights in a uniform distribution to have minimal accuracy drop after the quantization. The two studies~\cite{hoffman2019robust,shkolnik2020robust} are the most relevant to ours in that sharing the same objective of robust quantization. However, according to our observation, the former becomes unstable in advanced networks (i.e., MobileNet-V2/V3)  and the latter is orthogonal to ours. Our methods show results comparable to those of KURE, and we can maximize the endurance of the network by applying ours with KURE jointly, as will be shown in \cref{sec:exp}. 


\section{Motivation}\label{sec:motiv}
In this section, we provide the analysis of the error sources induced by quantization and explain the motivations to enhance the robustness of the networks for each error source. 

\begin{figure}[t]
\begin{subfigure}{.5\textwidth}
  \centering
  \includegraphics[width=1.\linewidth]{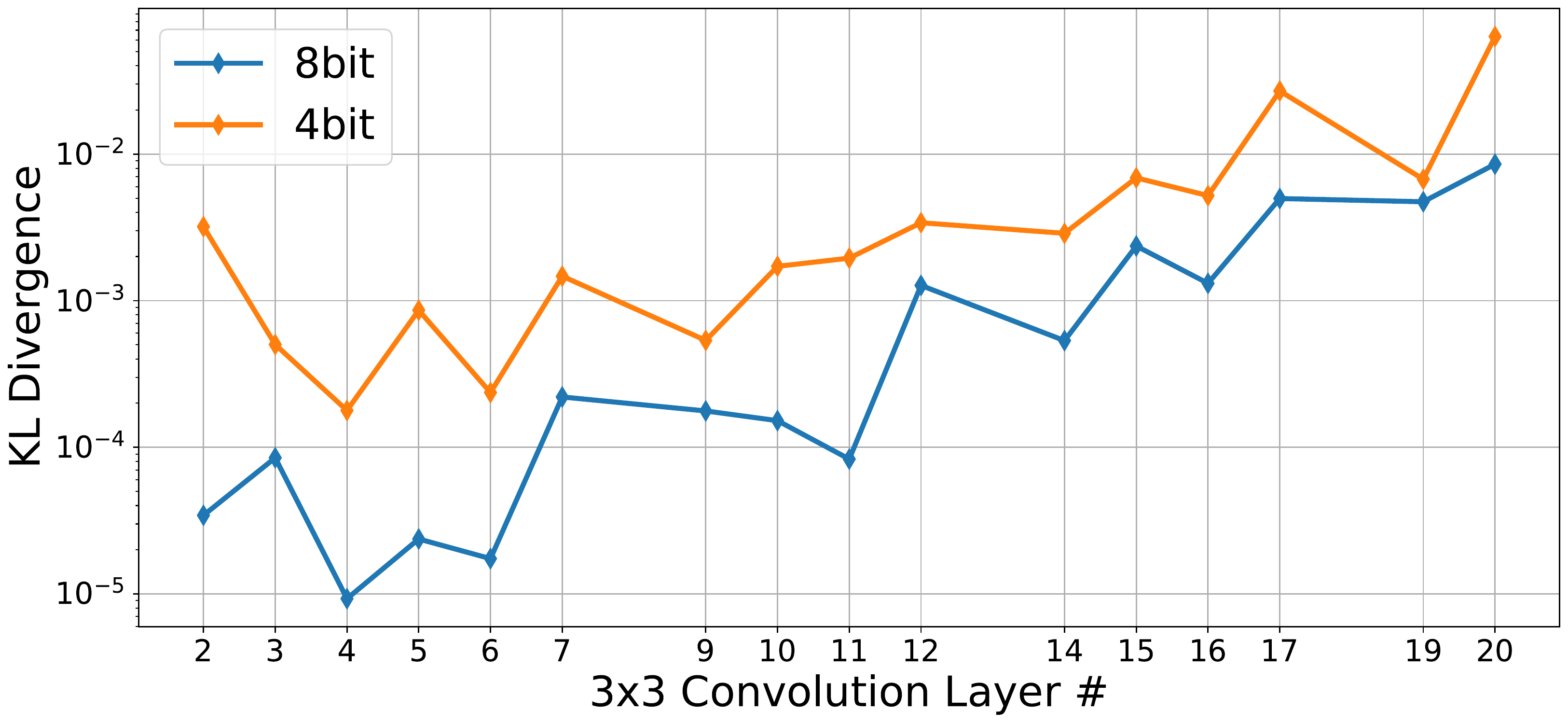}
  \caption{ResNet-18}
  \label{fig:propaerror_r18}
\end{subfigure}
\begin{subfigure}{.5\textwidth}
  \centering
  \includegraphics[width=1.\linewidth]{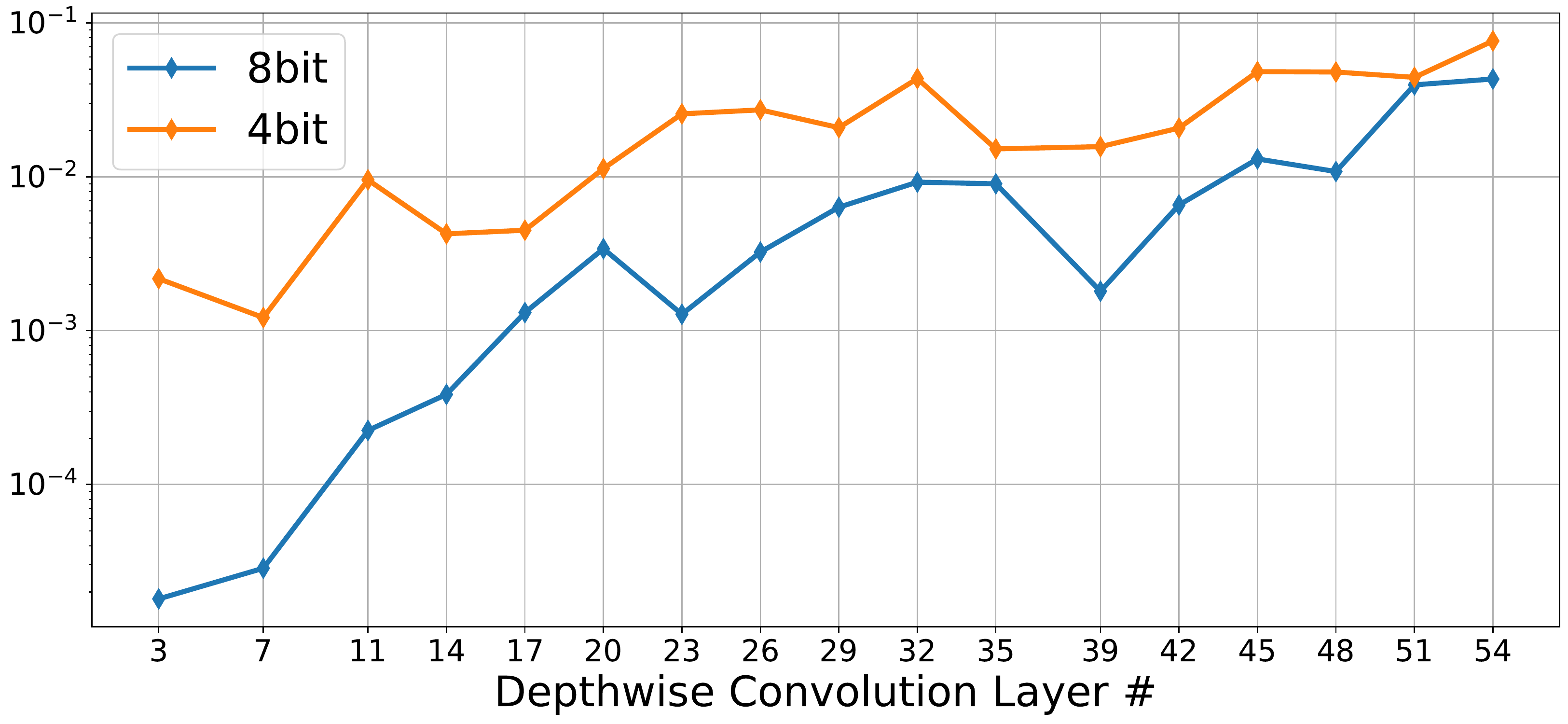}
  \caption{MobileNet-V2}
   \label{fig:propaerror_mv2}
\end{subfigure}
\caption{KL divergence of 3x3 convolution and depthwise convolution output between original model and model quantized with PTQ~\cite{banner2018aciq} (a) ResNet-18 (b) MobileNet-V2.}
\label{fig:propaerror}
\end{figure}

\subsection{Reduction of Error Propagation}\label{subsec:pe}
Several previous studies have indicated that quantization introduces the distortion of statistics compared to the original distribution~\cite{lin2018defensive,finkelstein2019fighting}. Moreover, when we apply quantization over the entire network, each layer introduces additional distortion to the output. As a result, the error continues to propagate and accumulate over the networks, as shown in \cref{fig:propaerror}, resulting in a large amount of accuracy degradation. Many studies related to PTQ have attempted to mitigate this problem by explicitly minimizing the difference in statistics before and after the quantization~\cite{finkelstein2019fighting,brock2021characterizing}. However, in this paper, we propose an alternative approach to minimize the difference in statistics on any quantization algorithms.

To achieve this goal, we focus on minimizing the biased quantization error problem~\cite{nagel2019data,brock2021characterizing}. Consider the linear or convolution operation $y = W\cdot x$, where $W$ is an arbitrary fixed weight and $x$ is an activation assumed i.i.d. variable with $E_i[x_i] = \mu_x$. In this condition, we can estimate the expected value of a single output unit $y_j=\Sigma_i^NW_{j,i}\cdot x_i$: 
\begin{equation}
    E_j[y_j] = N\mu_xE_i[W_{j,i}],
\end{equation}
where $N$ is the number of elements and $i$ is the index of input $x$. When we apply the quantization to the weight, the expected value of output drifts due to the distortion of the weight as follows:
\begin{equation}\label{eq:goal}
    E_j[\tilde{y_j} - y_j] = N\mu_x\cdot \Big(E_i\big[Q(W_{j,i})\big] - E_i\big[W_{j,i}\big]\Big),
\end{equation}
where $y_j$ and $\tilde{y_j}$ are the j-th output with the original weight and quantized weight, respectively. To minimize error propagation, we should minimize the difference of averaged weight in the output-channel dimension. However, satisfying \cref{eq:goal} strictly for any quantization algorithm is highly challenging. To ease the difficulty of the objective, we adopt the additional condition as given by:
\begin{equation}\label{eq:cond}
    E_i[Q(W_{j,i})] = E_i[W_{j,i}] = 0. 
\end{equation}
By forcing the mean of the weights before and after the quantization toward 0, \cref{eq:goal} is satisfied as a sufficient condition. When the full-precision weight is symmetric, the mean of the full-precision weight is zero. In addition, when we apply a symmetric quantization, which is commonly used for the weight quantization due to hardware compatibility~\cite{jacob2018integer,NVIDIA_8bit,IntArithmetic}, the drift in the positive values could be amortized by the drift in the negative values; thereby, the mean of the quantized weight is also zero. 

In summary, if we can force the full-precision weight in a symmetric distribution, the statistics distortion and error propagation after the quantization could be minimized. 
Unlike the explicit bias correction process~\cite{nagel2019data}, weight symmetry inherits the robustness against bias drift, enabling us effortless transition to different quantization policies. 

To guide the convergence of weight toward the symmetric distribution, we propose a novel regularization in \cref{sec:symreg}. 
When we apply this regularization during pre-train the model, the difference in statistics before and after PTQ is reduced. Furthermore, the statistics distortion is also minimized when we utilize the fine-tuned weight after QAT in different bit-widths without an additional fine-tuning stage, helping to maintain the quality of output.

\subsection{Range Clamping for Error Minimization}\label{sec:minerror}
\begin{figure*}[t!]
    \centering
    \begin{minipage}[t]{\dimexpr.5\textwidth-1em}
        \centering
        \includegraphics[trim= 1mm 10mm 0mm 0mm, width=1 \linewidth]{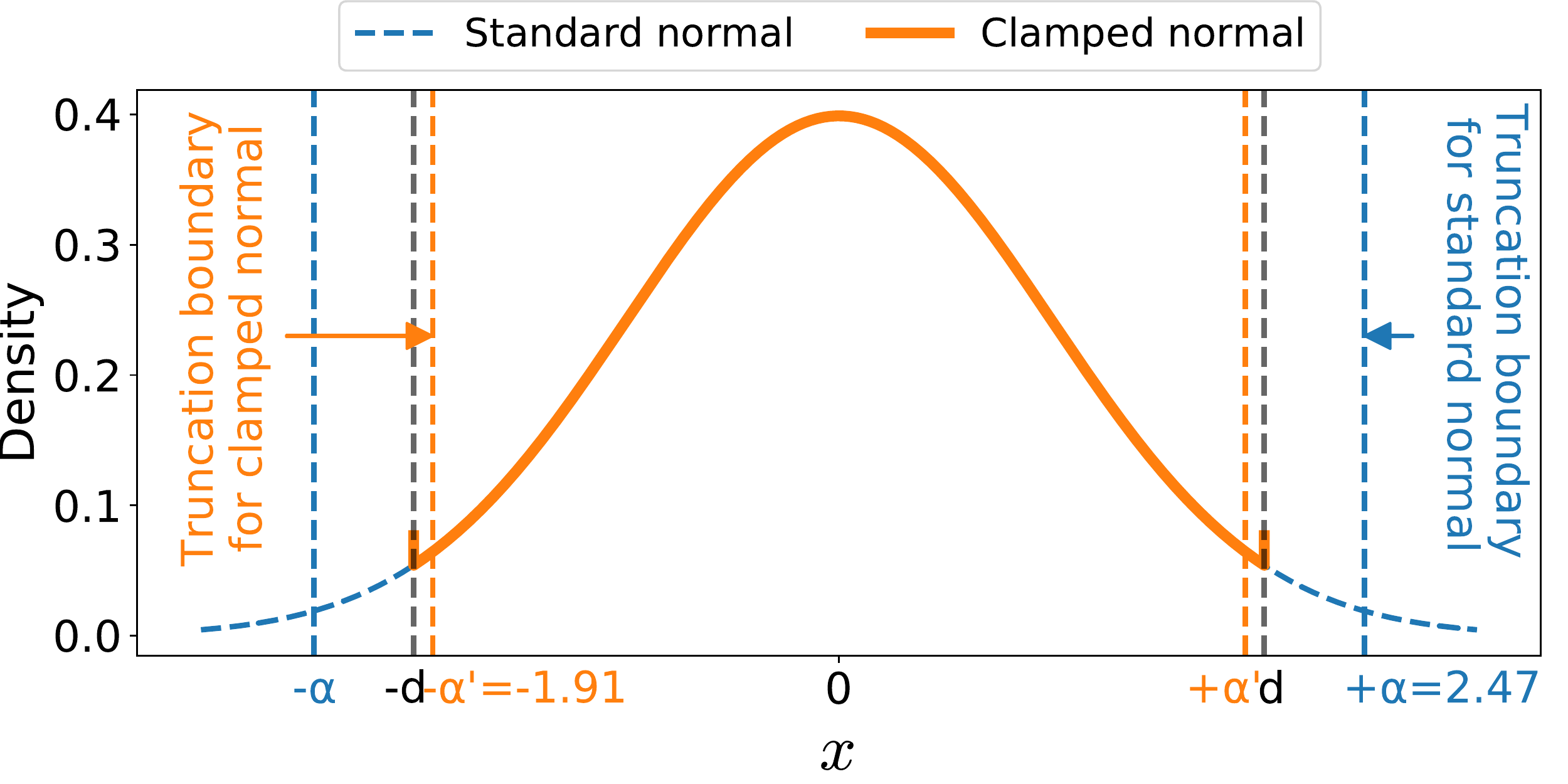}
        \caption{Histogram of standard normal vs clamped normal $(d = \pm 2)$, and their truncation boundaries.}
        \label{fig:normal_distribution}
    \end{minipage}\hfill
    \begin{minipage}[t]{\dimexpr.5\textwidth-1em}
        \centering
        \includegraphics[trim= 0mm 10mm 0mm 0mm, width=1\linewidth]{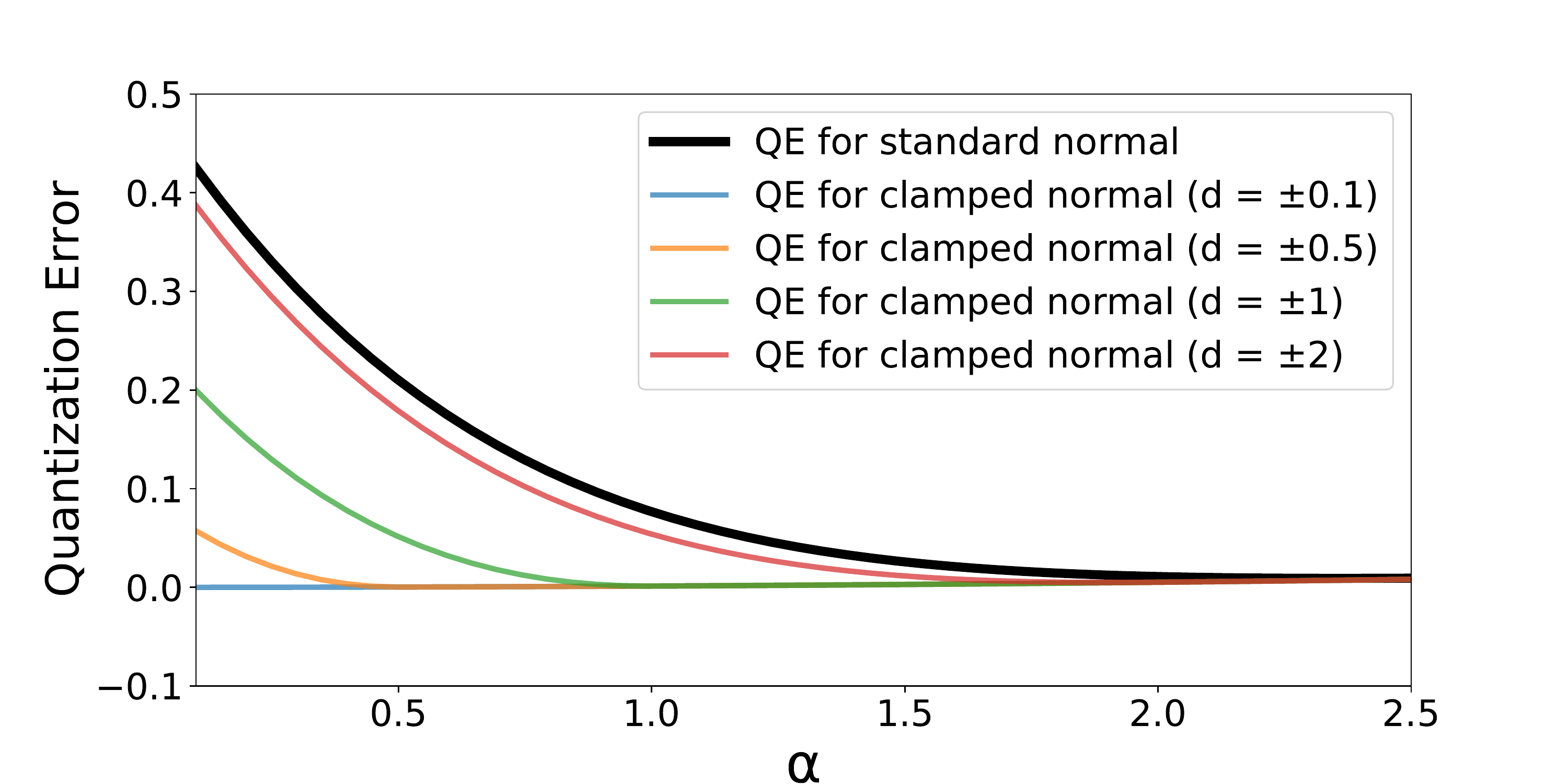}
       \caption{Analysis of quantization error for standard normal and clamped normal.}
       \label{fig:quantization_error}
    \end{minipage}
\end{figure*}

In linear quantization, the quantization levels are evenly distributed in between the truncation boundaries. Applying quantization to the full-precision tensor induces quantization error, which can be decomposed into the truncation error and the rounding error. The truncation and rounding errors are inevitable because of the limited number of quantization levels, but the difference of domain (i.e., the unbounded full-precision and the truncated quantization) enlarges the quantization error. A previous study~\cite{Outlier} pointed out that the data with infrequent but large values require a widened truncation boundary, which increases the rounding error significantly. To mitigate the quantization error, the domain of full-precision data should be narrowed to a bounded range.

Motivated by this limitation, we propose a straightforward idea of introducing the range clamping to the full-precision weight\footnote[1]{Please note that we intentionally use different expressions to distinguish quantization's truncation and the clamping of full-precision data.}, as shown in \cref{fig:normal_distribution}. Assume that the original weight follows a normal distribution whose PDF is $f(x)\sim N(0,1)$, just as the convention of previous studies~\cite{banner2018aciq,shkolnik2020robust}. When we apply the b-bit symmetric quantization toward minimizing the L2 norm, the quantization error can be estimated as follows~\cite{banner2018aciq}:
\begin{equation}
\begin{split}
&\text{Quantization Error} = E[(W - Q(W))^2] \\
&\approx \overbrace{2 \cdot \int_{\alpha}^{\infty} f(x)\cdot(x-\alpha)^2dx}^{\text{truncation error}} + \overbrace{\frac{\alpha^2}{3\cdot 2^{2b}}}^{\text{rounding error}},
\label{eq:ACIQ ERROR}
\end{split}
\end{equation}
where $\alpha$ is the truncation boundary that minimizes $||W - Q(W)||_2$. On the other hand, the clamping modifies the distribution of weight, whose cumulative distribution function $G(x)$ is expressed as 
\begin{equation}
\begin{split}
& G(x;d)  =
\begin{cases}
0,  &  x \leq -d \\
F(x) + F(-|d|), & -d < x < d \\
1,  &  d \leq x, \\
\end{cases}
\end{split}
\label{eq:d-clamped normal CDF}
\end{equation}
where $d$ is the newly introduced clamping target and $F(x)$ is the cumulative distribution function of $f(x)$. Then, the quantization error of the clamped distribution is expressed as
\begin{equation}
\begin{split}
&\text{Quantization Error}' = E[(W' - Q(W'))^2] \\
&\approx \overbrace{2\cdot\Big(F(-|d|)\cdot (d-\alpha')^2 + \int_{\alpha'}^{d} f(x)\cdot(x-\alpha')^2dx\Big)}^{\text{truncation error}}   + \overbrace{\frac{\alpha'^2}{3\cdot 2^{2b}}}^{\text{rounding error}}, 
\label{eq:ACIQ_clamped_normal_ERROR}
\end{split}
\end{equation}
where $\alpha'$ is the truncation boundary that minimizes $||W' - Q(W')||_2$.

When we compare the errors of \cref{eq:ACIQ ERROR} and \cref{eq:ACIQ_clamped_normal_ERROR}, \cref{eq:ACIQ_clamped_normal_ERROR} always has a smaller error than \cref{eq:ACIQ ERROR}, as shown in \cref{fig:quantization_error}. The proofs of \cref{eq:ACIQ ERROR} to \cref{eq:ACIQ_clamped_normal_ERROR} are provided in the supplementary material.
 
This analysis indicates that the range clamping of full-precision data is beneficial for minimizing quantization error. Thereby, if we train a network with the range clamping nonlinearity, the network could have a strong endurance for quantization. In addition, the quantization error could be minimized in the different precision, resulting in low accuracy loss other than the bit-width we quantized.  Indeed, a similar idea was addressed in MobileNet-V2~\cite{sandler2018mobilenetv2} in terms of ReLU6 for activation. The range clamping can be seen as an extension of the idea of ReLU6 for weight.
 
One possible drawback of this idea is the accuracy degradation due to the limited degree of freedom. For instance, when we set the clamping target close to zero, the quantization error becomes negligible, but the training from scratch may fail or show poor accuracy. Therefore, it is necessary to find a sweet spot that reduces the quantization error by a large margin while maintaining the quality of output. Empirically, we determine the practical implementation of the truncation that has negligible quality degradation while minimizing quantization loss significantly and provide it in \cref{sec:SatNL}.

\subsection{Inherited Robustness against Quantization}
\label{subsec:re}

\begin{figure*}[t!]
    \centering
    \begin{minipage}[t]{\dimexpr.675\textwidth-1em}
        \centering
        \includegraphics[trim= 0mm 5mm 0mm 0mm, width=1.0\linewidth]{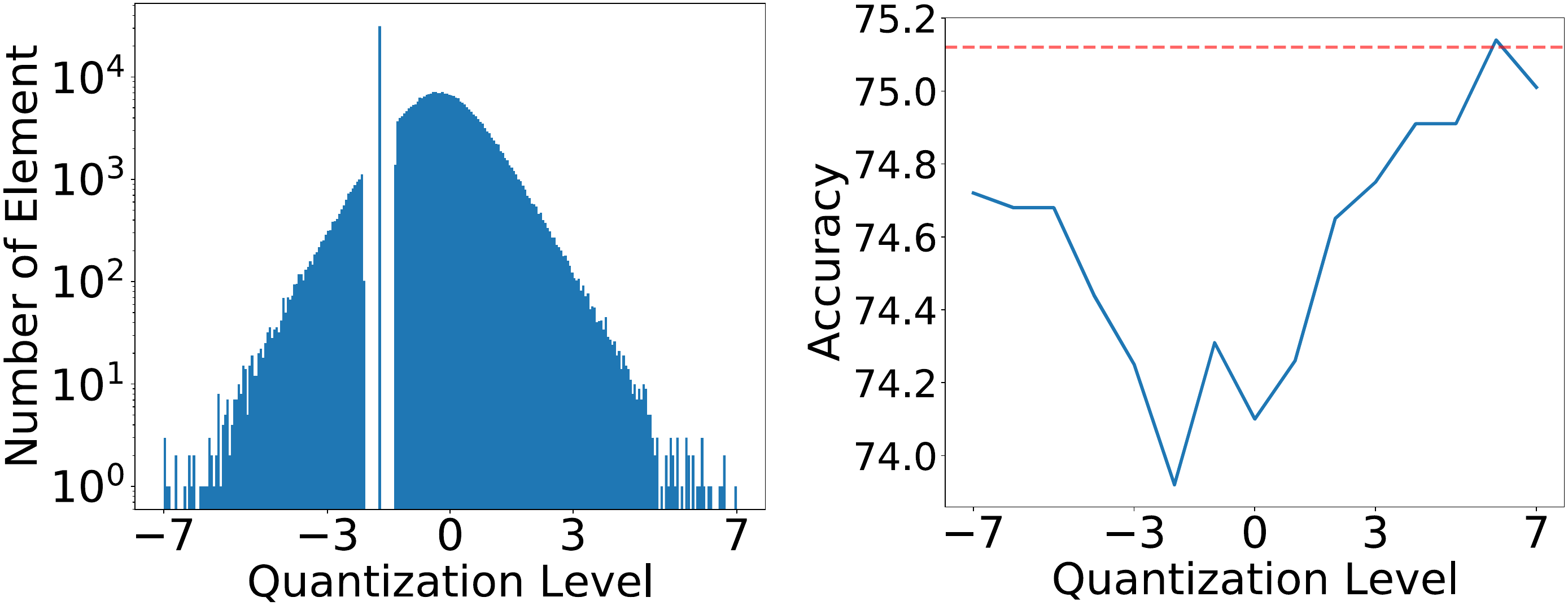}
        \caption{Example of single-level quantization and corresponding result. The dashed line represents the baseline full-precision accuracy, and the solid line shows that of the quantized network.}
        \label{fig:single level quantization}
    \end{minipage}\hfill
    \begin{minipage}[t]{\dimexpr.325\textwidth-1em}
    \centering
    \includegraphics[trim= 7mm 5mm -4mm 0mm, width=1.03\linewidth]{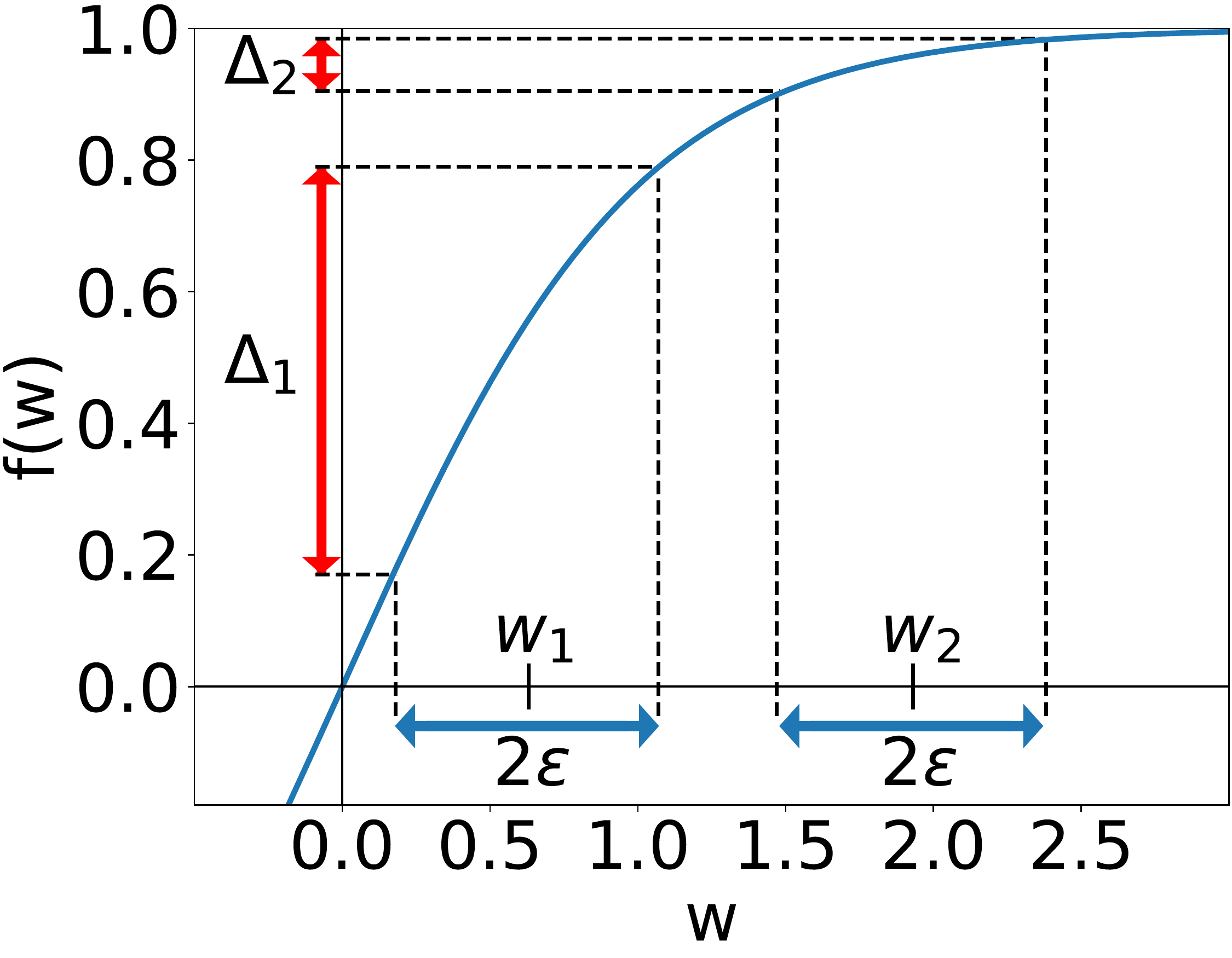}
   \caption{Saturating nonlinearity for weight and the adversarial boundary of ASAM.}
       \label{fig:tanh_asam}
    \end{minipage}
\end{figure*}

In addition to the range clamping for error minimization, we inspect another approach that enhances the inherited robustness of networks against quantization based on Hessian-aware loss sharpness minimization. Recently, several studies have focused on enhancing the generalization of neural networks based on the training pipeline aware of the sharpness of loss surface~\cite{foret2020sam,kwon2021asam}. Those studies have aimed to guide the convergence of networks toward the flat minimum having smaller Hessian values, where the minor distortion of weight could be ignored without affecting the output. From a similar perspective, the Hessian of weight is utilized as an important metric for measuring the sensitivity of networks regarding quantization~\cite{dong2019hawq,dong2019hawqv2,li2021brecq}. 
Motivated by these examples, we try to adopt the Hessian-aware training to enhance the robustness of networks for quantization by guiding the convergence of networks into a smooth loss surface. We utilize (A)SAM~\cite{foret2020sam,kwon2021asam}, and empirically observe that training with (A)SAM is beneficial for minimizing quantization error.

However, we also observe that there is room for improvement in terms of enhancing robustness for quantization with (A)SAM, because the quantization sensitivity differs depending on the value of quantization levels. \cref{fig:single level quantization} shows the accuracy degradation after applying single-level PTQ to the weights of MobileNet-V2 on CIFAR-100, where the weights corresponding to the specific quantization level are quantized while the rest remain in full-precision. As shown in the figure, the accuracy degradation increases when the smaller weights are quantized. This indicates that the weights near zero are more vulnerable to quantization than the weights having large values. We speculate that because the majority of weights are concentrated near zero, the accumulated error of quantization is inversely proportional to the magnitude of the value. Thereby, to maintain the quality of output with quantization, the sharpness minimization should be applied in different strengths depending on the magnitude of weight. 

To realize the aforementioned objective, we propose a novel idea that introduces the saturating nonlinearity to the weight, as visualized in \cref{fig:tanh_asam}. When we apply (A)SAM with the nonlinearity, the robustness boundary of (A)SAM, which is equally assigned in the weight before the nonlinearity, covers a different range in output depending on the slope of the nonlinearity. SatNL has the largest slope near zero; as a result, the effect of the (A)SAM algorithm turns friendly to the quantization.

\section{Implementation}\label{sec:imple}

Based on the motivations in the previous section, we propose the practical implementations, called symmetry regularization (SymReg) and saturating nonlinearity (SatNL). 
 
\subsection{Symmetry Regularization}\label{sec:symreg}

\begin{figure}[t]
   \centering
   \includegraphics[width=0.7\linewidth]{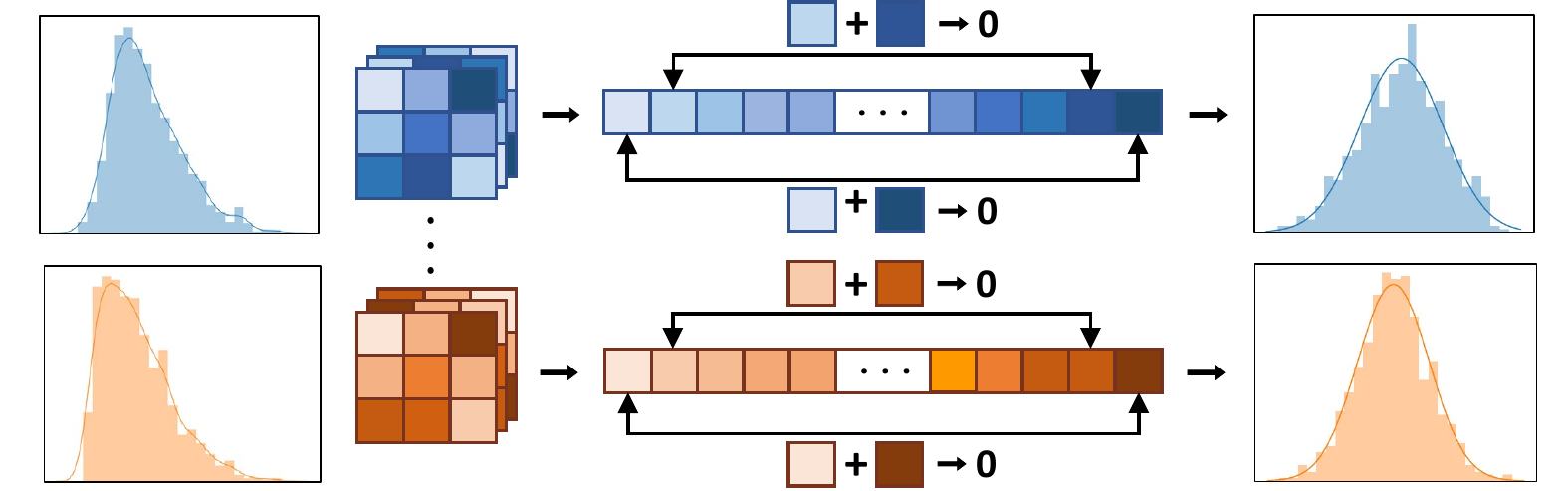}
   \caption{Visualization of proposed SymReg.}
   \label{fig:symlosseffect}
   \hfill
\end{figure}
\vspace{-2mm}

In \cref{subsec:pe}, we claimed that the symmetric weight could minimize the error propagation. To realize this, we propose an additional regularization called symmetry regularization (SymReg). 

The weight symmetry is achievable when every weight has a corresponding mate with an identical magnitude but a different sign. After sorting the weights and assigning the index in ascending order, we can make pairs where one element is selected in ascending order $\tilde{w}_i$ while the other is in descending order $\tilde{w}_{N-i}$. When we calculate the $L1$ norm of the sum of each pair, the expectation should become zero as follows:
\begin{equation}
\frac{2}{N} \sum_{n=1}^{\frac{N}{2}}|\tilde{w}_n + \tilde{w}_{N-n}| = 0.
\label{eq:sym_property}
\end{equation}

Based on this intuition, we design a layer-wise SymReg that guides the convergence of weight into the symmetric distribution as shown in \cref{fig:symlosseffect}. The SymReg is defined as:
\begin{equation}
\mathcal{L}_{sym1} = \frac{2}{C\cdot N}\sum_{c=1}^{C}\sum_{i=1}^{\frac{N}{2}}|w^c_i + w^c_{N-i}|,
\label{eq:sym_loss1}
\end{equation}
where $w^c_i$ represents the i-th smallest element in the c-th channel.  

$L_{sym1}$ is applied when we train the full-precision network in addition to the conventional loss functions. However, if the restriction of $L_{sym1}$ is too severe, it could lead to a minor accuracy degradation. To minimize it, we propose the relaxed SymReg, which measures the symmetry in a 2:2 relation with more degree of freedom instead of a 1:1 relation. We empirically observe that more than 3:3 relation degrades the benefit of SymReg.

\begin{equation}
\mathcal{L}_{sym2} = \frac{4}{C\cdot N}\sum_{c=1}^{C}\sum_{i=1}^{\frac{N}{4}}|w^c_{2i} + w^c_{2i+1} + w^c_{N-2i} + w^c_{N-2i-1}|.
\label{eq:sym_loss2}
\end{equation}
In the experiment, we combine $\mathcal{L}_{sym1}$ and $\mathcal{L}_{sym2}$ adequately to minimize accuracy degradation while exploiting the benefit of propagation error minimization. The overall loss is expressed as follows: 
\begin{equation}
\mathcal{L} = \mathcal{L}_{CE} + \lambda_1 \cdot \mathcal{L}_{sym1} + \lambda_2 \cdot \mathcal{L}_{sym2}.
\end{equation}

\subsection{Saturating Nonlinearity (SatNL)}\label{sec:SatNL}

In \cref{sec:minerror}, we showed that the truncation of full-precision weight could be highly beneficial for minimizing quantization error. In addition, in \cref{subsec:re}, we showed the empirical analysis indicating that the quantization sensitivity differs depending on the magnitude of the weight. To resolve these two problems simultaneously, we propose applying a specialized nonlinearity function $f$ on top of the weight as follows: $conv(W, X) \xrightarrow{} conv(f(W), X).$

The nonlinearity should satisfy three properties.
1. It needs to be an odd function. Because we assume that the weight is quantized by the symmetric quantization, it would be better to have an identical range of the negative and positive regions to maximize the quantization level efficiency. 
2. The range of output needs to be bounded. To satisfy the criteria of \cref{sec:minerror}, the weight after the nonlinearity should be narrowed to a certain range.
3. The slope is gradually decreased as the input value is increased. To maximize the benefit of (A)SAM, the nonlinearity should be saturated as the value is increased. 
Empirically, we choose the hyper-tangent (tanh) function as nonlinearity which satisfies all three conditions. Note that the normalized tanh was used in QAT studies~\cite{ZhoWu16Dorefa}, but there is a fundamental difference in terms of the intention of using it. 
In this study, we exploit the tanh function intentionally to maximize the robustness against quantization and turn (A)SAM algorithm friendly to quantization. Other nonlinearity functions could be applicable when the three conditions are met. According to our experiments, the impact on the final accuracy is negligible regardless of the nonlinearity functions, while the desired properties for robustness are valid. Additional analysis is in the supplementary material. 

\section{Experiments}\label{sec:exp}
To show the superiority of the proposed methods, we conduct extensive studies on CIFAR-100 and ImageNet datasets with representative networks (i.e., ResNet-18~\cite{ResNet}, MobileNet-V2~\cite{sandler2018mobilenetv2}, and MobileNet-V3~\cite{howard2019mobilenetv3}). We use layer-wise asymmetric quantization for activation and output-channel-wise symmetric quantization for weight to enable acceleration on existing hardwares~\cite{jacob2018integer,wu2020integer,IntArithmetic,NVIDIA_8bit}.

For QAT, we apply LSQ~\cite{Esser2020lsq}, which is the advanced differentiable quantization scheme that allows the quantization of ResNet-18 into 3-bit without accuracy loss in the ImageNet. For PTQ, we apply the ACIQ~\cite{banner2018aciq}, AdaQuant~\cite{AdaQuant}, and QDrop~\cite{qdrop} algorithms. ACIQ is a well-known PTQ that analytically finds the optimal quantization boundary, and QDrop is a state-of-the-art PTQ algorithm with small calibration sets.
By utilizing multiple quantization algorithms with different properties, we aim to validate the universal robustness.

Moreover, we use a common practice that fixes the bit-width of the first and the last layers into 8-bit. All of the other layers are quantized into the given bit-width identically unless explicitly specified otherwise. SymReg is not applied to the depthwise convolution and the linear layer in MobileNet-V2/V3.
In 3$\times$3 depthwise convolution, SymReg degrades the expression capability significantly by forcing one out of nine elements to become zero. In ASAM, $\rho$ is fixed as 1 except 0.2 for MobileNet-V3 because ASAM with high $\rho$ becomes unstable. The hyper-parameters of SymReg $\lambda_1/ \lambda_2$ are empirically set as 0.1/0.1 for ImageNet and CIFAR-100. The details of the training parameters (i.e., learning rate, decay, epochs, etc.), are provided in the supplementary material. 

\subsection{Robustness of Bit-Precision for PTQ}

\begin{table*}[t!]
\caption{Results of applying PTQ to baseline and network with proposed ideas on the ImageNet dataset. The values in the table represent the top-1 accuracy. The dashed cells represent the points where the PTQ fails to converge.}
\centering
\resizebox{\hsize}{!}{
\begin{tabular}{ccccccccccc}
\hline
\multirow{2}{*}{Model}                            & \multirow{2}{*}{PTQ}                           & \multirow{2}{*}{Method}       & \multicolumn{8}{c}{Weight/activation bit-width configuration}                             \\ \cline{4-11} 
                                                  &                                                &                               & FP    & 4/FP   & 3/FP   & 2/FP                        & 6/6    & 5/5    & 4/4    & 3/3    \\

\hline

\multicolumn{1}{c|}{\multirow{6}{*}{ResNet-18}}   & \multicolumn{1}{c|}{\multirow{2}{*}{ACIQ\cite{banner2018aciq}}}     & \multicolumn{1}{c|}{Baseline} & 70.54 & 47.44  & -      & \multicolumn{1}{c|}{-}      & 68.70   & 64.87 & 38.46 & -      \\
\multicolumn{1}{c|}{}                             & \multicolumn{1}{c|}{}                          & \multicolumn{1}{c|}{Ours}     & 70.92 & 69.22 & 49.06 & \multicolumn{1}{c|}{-}      & 70.02  & 68.99 & 66.65 & 42.95 \\

\cline{2-11} 
\multicolumn{1}{c|}{}                             & \multicolumn{1}{c|}{\multirow{2}{*}{AdaQuant\cite{AdaQuant}}} & \multicolumn{1}{c|}{Baseline} & 70.54 & 69.29 & 66.18 & \multicolumn{1}{c|}{3.23}  & 70.17 & 69.55 & 67.67 & 57.57 \\
\multicolumn{1}{c|}{}                             & \multicolumn{1}{c|}{}                          & \multicolumn{1}{c|}{Ours}     & 70.92 & 70.36 & 68.84 & \multicolumn{1}{c|}{48.39} & 70.75 & 70.37  & 69.35 & 64.04 \\

\cline{2-11} 
\multicolumn{1}{c|}{}                             & \multicolumn{1}{c|}{\multirow{2}{*}{QDrop\cite{qdrop}}} & \multicolumn{1}{c|}{Baseline} &  70.54 & 70.15 & 69.39 & \multicolumn{1}{c|}{66.40}  & 70.27 & 69.93 & 68.91 & 65.75 \\
\multicolumn{1}{c|}{}                             & \multicolumn{1}{c|}{}                          & \multicolumn{1}{c|}{Ours}     & 70.92 & 70.69 & 70.06 & \multicolumn{1}{c|}{66.95} & 70.81 & 70.57 & 69.93 & 67.45  \\

\hline

\multicolumn{1}{c|}{\multirow{6}{*}{MobileNet-V2}} & \multicolumn{1}{c|}{\multirow{2}{*}{ACIQ\cite{banner2018aciq}}}     & \multicolumn{1}{c|}{Baseline} & 72.22 & 28.68 & -      & \multicolumn{1}{c|}{-}      & 69.30 & 64.20 & 18.15      & -      \\
\multicolumn{1}{c|}{}                             & \multicolumn{1}{c|}{}                          & \multicolumn{1}{c|}{Ours}     & 72.87 & 70.07 & 40.79 & \multicolumn{1}{c|}{-}      & 71.07 & 68.66 & 58.30 & 6.25      \\

\cline{2-11}
\multicolumn{1}{c|}{}                             & \multicolumn{1}{c|}{\multirow{2}{*}{AdaQuant\cite{AdaQuant}}} & \multicolumn{1}{c|}{Baseline} & 72.22 & 70.67 & 59.80 & \multicolumn{1}{c|}{-}      & 71.52 & 70.72 & 63.70 & -      \\
\multicolumn{1}{c|}{}                             & \multicolumn{1}{c|}{}                          & \multicolumn{1}{c|}{Ours}     & 72.87 & 72.23 & 69.03  & \multicolumn{1}{c|}{-}      & 72.27 & 71.76 & 68.91 & 18.36 \\

\cline{2-11} 
\multicolumn{1}{c|}{}                             & \multicolumn{1}{c|}{\multirow{2}{*}{QDrop\cite{qdrop}}} & \multicolumn{1}{c|}{Baseline} & 72.22 & 71.41 & 68.32 & \multicolumn{1}{c|}{48.68}  & 71.57 & 70.64 & 67.08 & 50.79 \\
\multicolumn{1}{c|}{}                             & \multicolumn{1}{c|}{}                          & \multicolumn{1}{c|}{Ours}     & 72.87 & 72.44 & 71.18 & \multicolumn{1}{c|}{61.68} & 72.61 & 72.05 & 69.87 & 62.55 \\

\hline

\multicolumn{1}{c|}{\multirow{4}{*}{MobileNet-V3}} & \multicolumn{1}{c|}{\multirow{2}{*}{ACIQ\cite{banner2018aciq}}}     & \multicolumn{1}{c|}{Baseline} & 74.52 & 29.65      & -      & \multicolumn{1}{c|}{-}      & -      & -      & -      & -      \\
\multicolumn{1}{c|}{}                             & \multicolumn{1}{c|}{}                          & \multicolumn{1}{c|}{Ours}     & 74.43 & 61.95      & 1.04      & \multicolumn{1}{c|}{-}      & -      & -      & -      & -      \\

\cline{2-11} 
\multicolumn{1}{c|}{}                             & \multicolumn{1}{c|}{\multirow{2}{*}{AdaQuant\cite{AdaQuant}}} & \multicolumn{1}{c|}{Baseline} & 74.52 & 72.92 & 64.17 & \multicolumn{1}{c|}{-}      & 72.73 & 68.95 & 43.88 & -      \\
\multicolumn{1}{c|}{}                             & \multicolumn{1}{c|}{}                          & \multicolumn{1}{c|}{Ours}     & 74.43 & 73.51 & 70.50 & \multicolumn{1}{c|}{2.87}      & 72.69 & 71.02 & 62.73 & -      \\


\hline
\end{tabular}
}
\label{tab:mainresult}
\end{table*}
Table \ref{tab:mainresult} shows the accuracy degradation after PTQ of ResNet-18, and MobileNet-V2/V3 on the ImageNet dataset. 
Our/baseline models are trained from scratch with/without the proposed ideas and then quantized into low-precision with PTQ algorithms. Our experiments show that the proposed methods are beneficial for minimizing the accuracy degradation after PTQ in every point regardless of the PTQ details. When we combine ours with the advanced PTQ (i.e., QDrop), we can quantize ResNet-18 into 4-bit with accuracy loss of less than 1~\%. Compared to the baseline, we reduce 1.02~\% of the top-1 accuracy degradation. In addition, the synergy of the advanced PTQ algorithm and our methods enables sub-8-bit quantization for the advanced networks with minimal accuracy degradation. QDrop with ours achieves 69.87~\% in 4-bit MobileNet-V2, which is the highest accuracy after 4-bit PTQ to the best of our knowledge. Notably, when we select the layer-wise bit-width depending on the sensitivity of layers, where the depthwise and squeeze-excitation layers are quantized into 8-bit and the rest of the layers are quantized into 4-bit with AdaQuant, MobileNet-V2/V3 shows 2.96~\%/4.34~\% of accuracy degradation respectively\footnote[2]{Note that the column of mixed-precision results is omitted in Table \ref{tab:mainresult} for brevity.}. In this configuration, we can enjoy the benefit of 4-bit computation in 85.3~\% of computation in the case of MobileNet-V2. Without ours, the accuracy degradation is 7.25~\% and 9.44~\% respectively in mixed precision, which is too poor to be used in real applications. The combination of network robustness enhancement and sensitivity-aware quantization could be a good candidate for practical deployment. According to our experiment, SymReg and SatNL extended the training time by 2.53~\% when we train MobileNet-V2 with ASAM on the ImageNet dataset. By spending this one-time overhead, a robust network that can minimize accuracy degradation regardless of the PTQ scheme can be achieved. Additional experiments compared with KURE are supported in supplementary materials.

\subsection{Ablation Study}
\begin{table}[!t]
\caption{Results of ablation study of proposed methods on MobileNet-V2 at CIFAR-100 dataset. The weight is quantized into the given bit-width with ACIQ. "All" means every method (+ SymReg + SatNL + ASAM). The values in the table represent the mean and standard deviation of top-1 accuracy with eight trials (10 trials except min. and max. results). FP means full-precision.}
 \centering
 \small
 \begin{tabular}{lc|c|c|c}
 \hline
 \multicolumn{1}{c}{}    & FP    & 4-bit & 3-bit & 2-bit \\ \hline
 Baseline                & 74.70{\tiny $\pm0.16$}&73.32{\tiny $\pm0.32$}&66.81{\tiny $\pm0.86$}&6.92{\tiny $\pm2.68$} \\
 + SymReg                & 74.66{\tiny $\pm0.16$}&73.49{\tiny $\pm0.30$}&69.69{\tiny $\pm0.81$}&25.01{\tiny $\pm4.80$} \\
 + SatNL                 & 74.80{\tiny $\pm0.11$}&73.43{\tiny $\pm0.36$}&68.65{\tiny $\pm1.23$}&14.45{\tiny $\pm2.93$} \\
 + SymReg + SatNL        & 74.41{\tiny $\pm0.10$}&73.34{\tiny $\pm0.31$}&69.66{\tiny $\pm0.62$}&33.68{\tiny $\pm5.87$} \\
 + ASAM                  & 75.52{\tiny $\pm0.19$}&74.42{\tiny $\pm0.16$}&69.71{\tiny $\pm0.71$}&11.80{\tiny $\pm3.96$}  \\
 + SatNL + ASAM          & 75.55{\tiny $\pm0.21$}&74.53{\tiny $\pm0.21$}&70.78{\tiny $\pm0.81$}&25.77{\tiny $\pm4.03$} \\
 + All                   & 75.33{\tiny $\pm0.16$}&74.55{\tiny $\pm0.27$}&72.17{\tiny $\pm0.22$}&39.27{\tiny $\pm2.56$} \\ \hline
 + KURE                  & 74.97{\tiny $\pm0.11$}&74.22{\tiny $\pm0.10$}&71.26{\tiny $\pm0.47$}&34.90{\tiny $\pm4.51$} \\
 + KURE + ASAM           & 75.57{\tiny $\pm0.18$}&74.96{\tiny $\pm0.11$}&72.48{\tiny $\pm0.46$}&42.73{\tiny $\pm4.71$} \\
 + KURE + All            & 75.41{\tiny $\pm0.3$}&74.91{\tiny $\pm0.29$}&73.34{\tiny $\pm0.25$}&37.58{\tiny $\pm2.99$}  \\

\hline
\end{tabular}
\label{tab:ablation}
\end{table}


Table \ref{tab:ablation} shows the effect of the proposed methods based on the accuracy degradation with PTQ. When we add an additional component of the proposed methods progressively (+~SymReg, +~SatNL, +~All), the PTQ error
is gradually reduced, showing that the proposed methods enhance the robustness of the network for PTQ in diverse aspects. SymReg and SatNL are beneficial for robustness but introduces a slight accuracy degradation in full-precision. However, the accuracy degradation could be mitigated with the assistance of ASAM. When we compare + ASAM and + SatNL + ASAM, the latter shows higher accuracy in full-precision and better robustness in lower precision, showing that the benefit of ASAM is boosted with SatNL.

When we compare the performance of the proposed methods with KURE, our methods lower accuracy degradation in the given bit-width than that of KURE. Meanwhile, when we apply ASAM with KURE, the robustness becomes comparable to ours. Moreover, the implementation details of KURE and ours are orthogonal. Thus, we can maximize the robustness by combining KURE and ours (KURE + ALL). This could be a state-of-the-art method for enhancing the robustness of networks, as far as we know.

\subsection{Robustness for Quantization Step Size }


\begin{figure}[t]
\begin{subfigure}{.26\textwidth}
    \centering
    \includegraphics[trim= 0mm 5mm 0mm 5mm, width=1.02\linewidth]{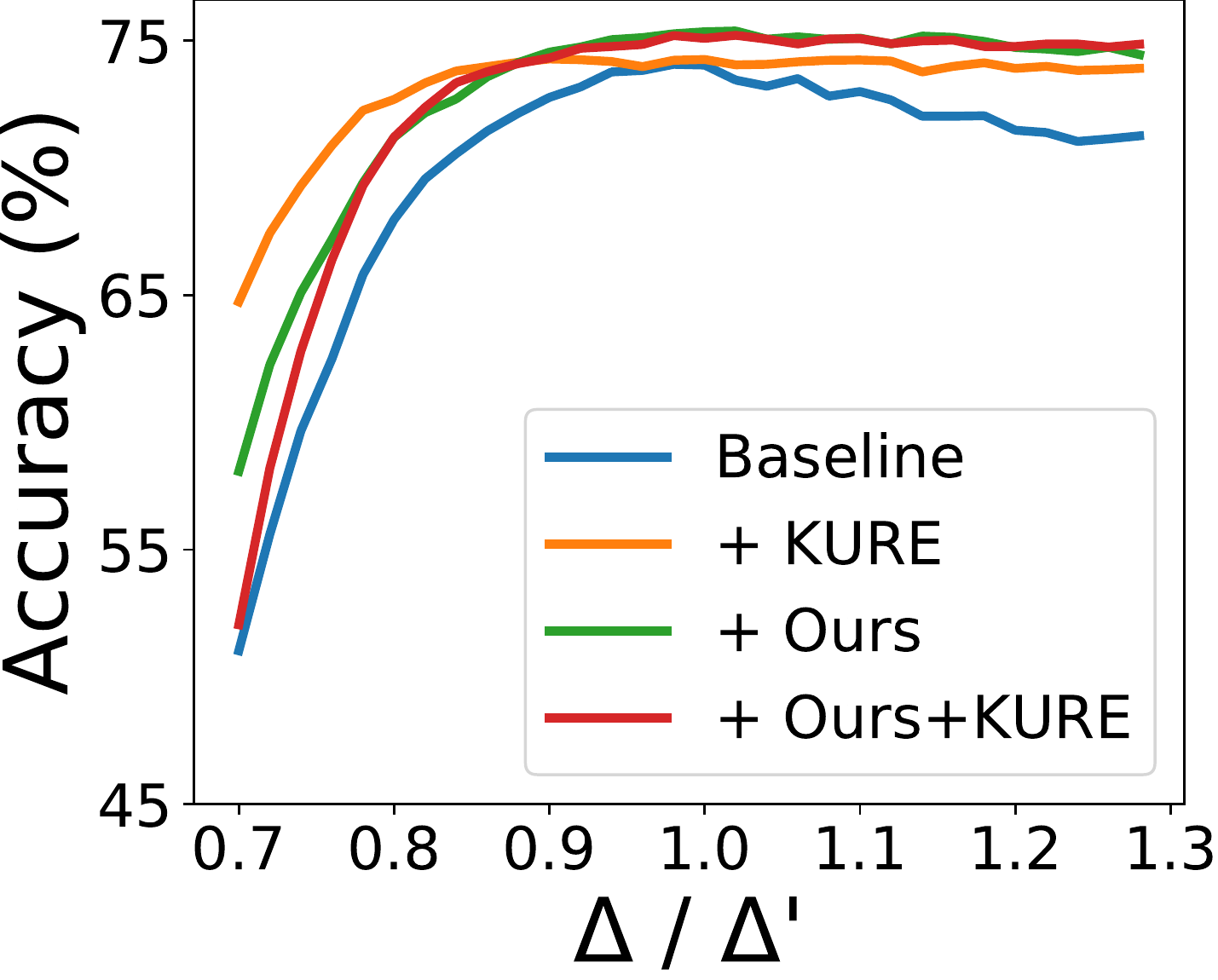}
  \captionsetup{justification=centering}
    \caption{ResNet-18\\PTQ, CIFAR-100}
\end{subfigure}
\begin{subfigure}{.24\textwidth}
    \centering
  \includegraphics[trim= 0mm 5mm 0mm 5mm, width=1.0\linewidth]{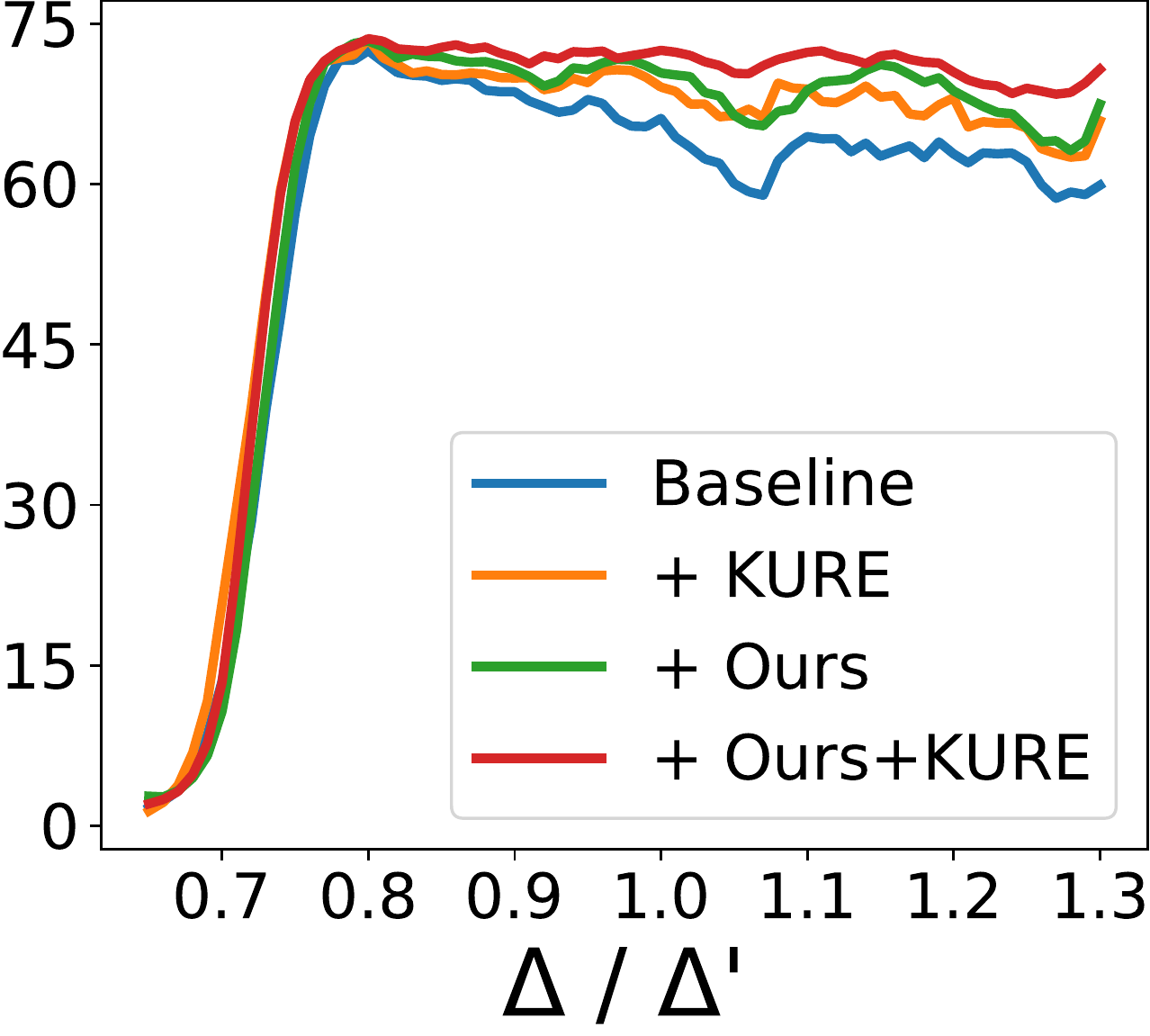}
  \captionsetup{justification=centering}
  \caption{MobileNet-V2\\PTQ, CIFAR-100}
\end{subfigure}
\begin{subfigure}{.24\textwidth}
    \centering
  \includegraphics[trim= 0mm 5mm 0mm 5mm, width=1.0\linewidth]{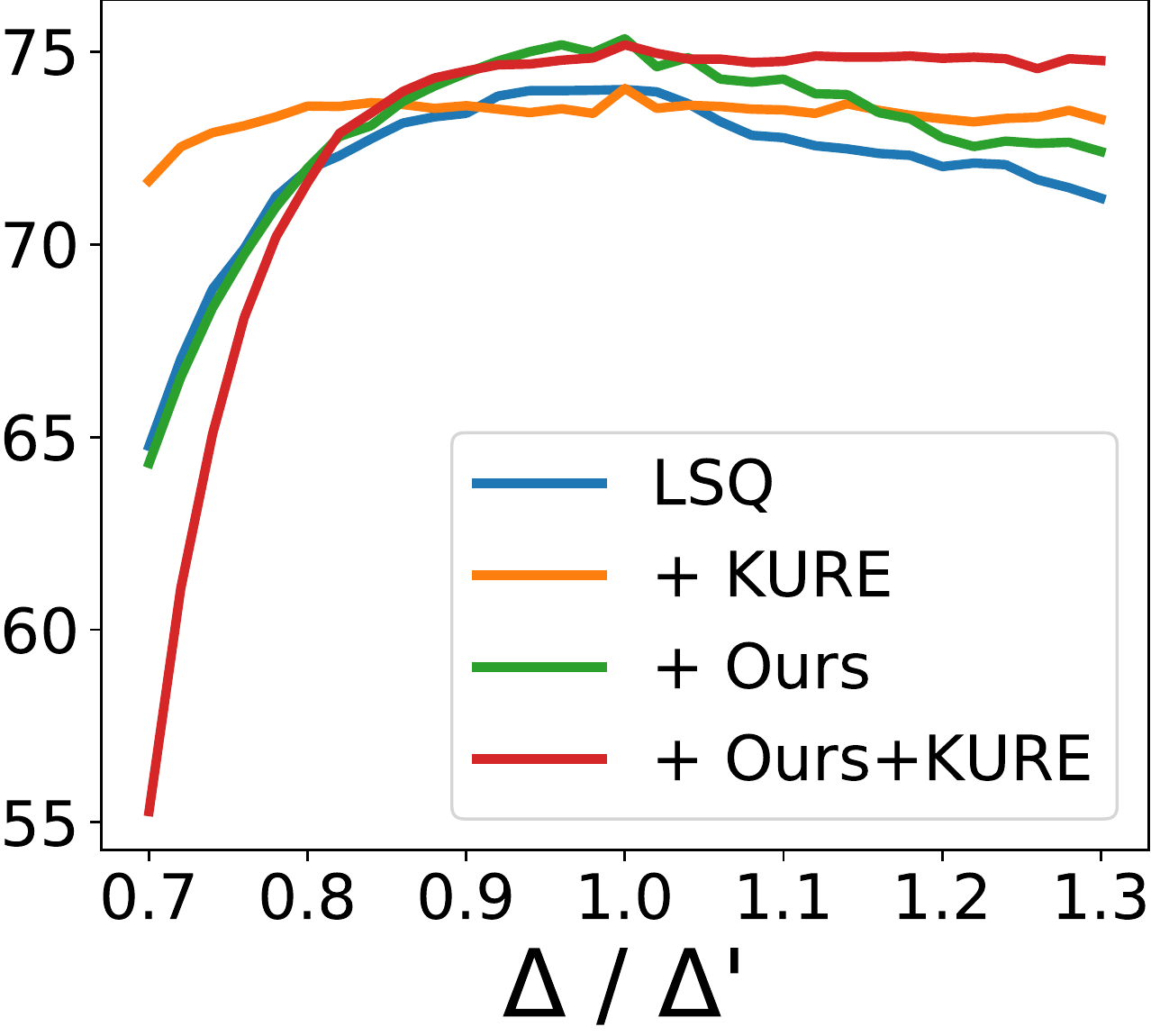}
  \captionsetup{justification=centering}
      \caption{ResNet-18\\QAT, CIFAR-100}
\end{subfigure}
\begin{subfigure}{.24\textwidth}
    \centering
  \includegraphics[trim= 0mm 5mm 0mm 5mm, width=1.0\linewidth]{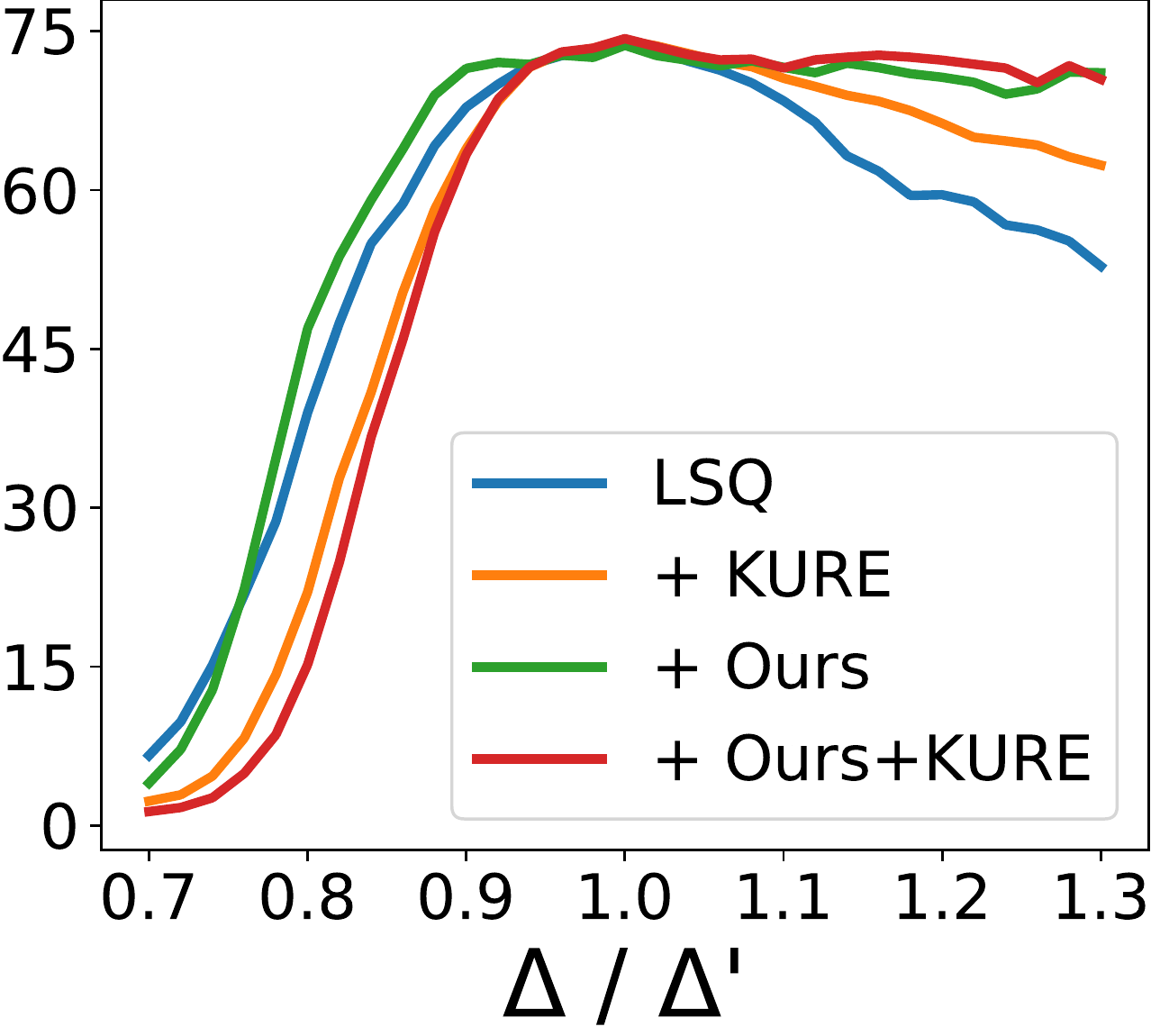}
  \captionsetup{justification=centering}
    \caption{MobileNet-V2\\QAT, CIFAR-100}
\end{subfigure}

\begin{subfigure}{.26\textwidth}
\centering
  \includegraphics[trim= 0mm 5mm 0mm 0mm, width=1.0\linewidth]{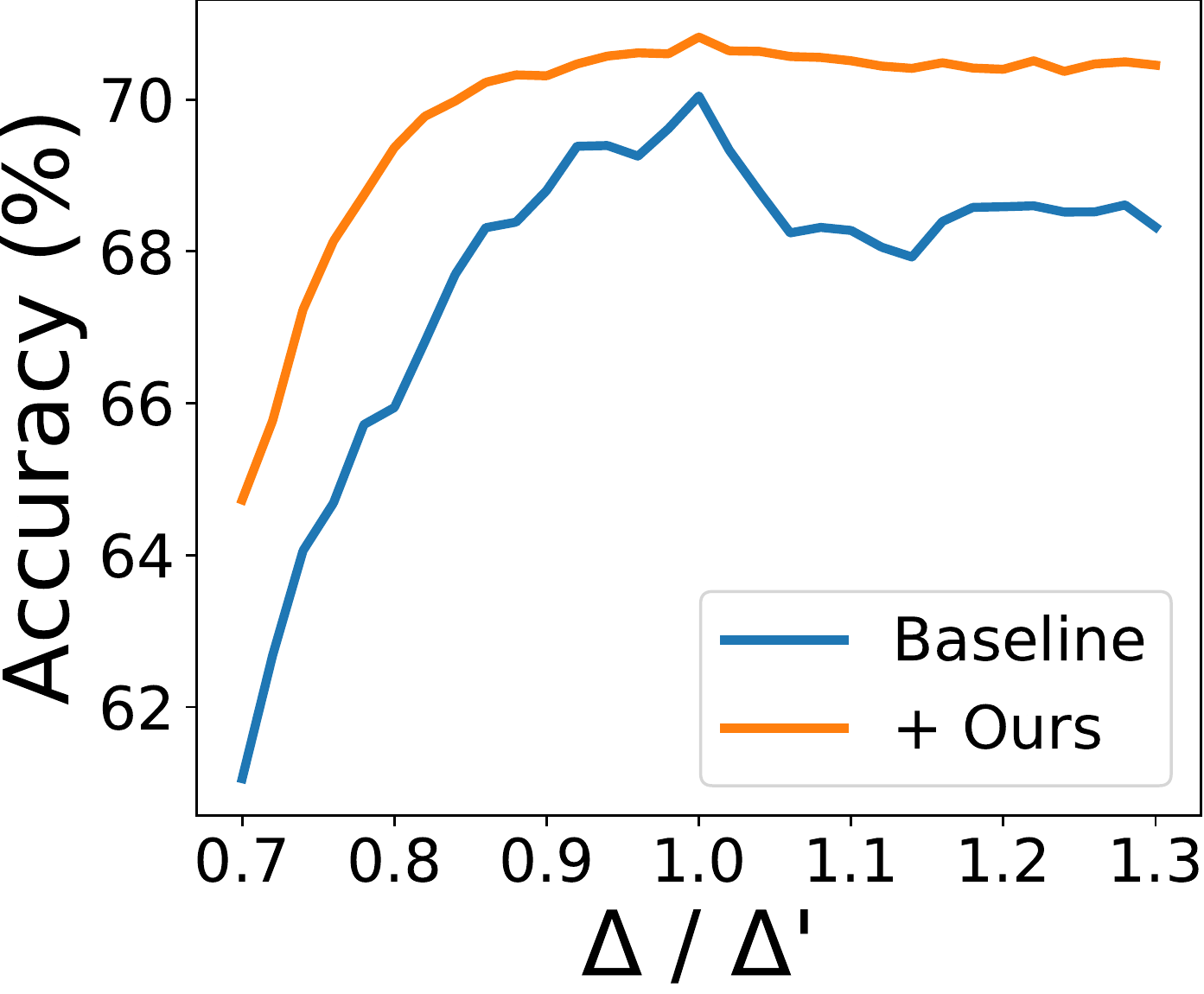}
  \captionsetup{justification=centering}
  \caption{ResNet-18\\PTQ, ImageNet}
\end{subfigure}
\begin{subfigure}{.24\textwidth}
\centering
  \includegraphics[trim= 0mm 5mm 0mm 0mm, width=1.0\linewidth]{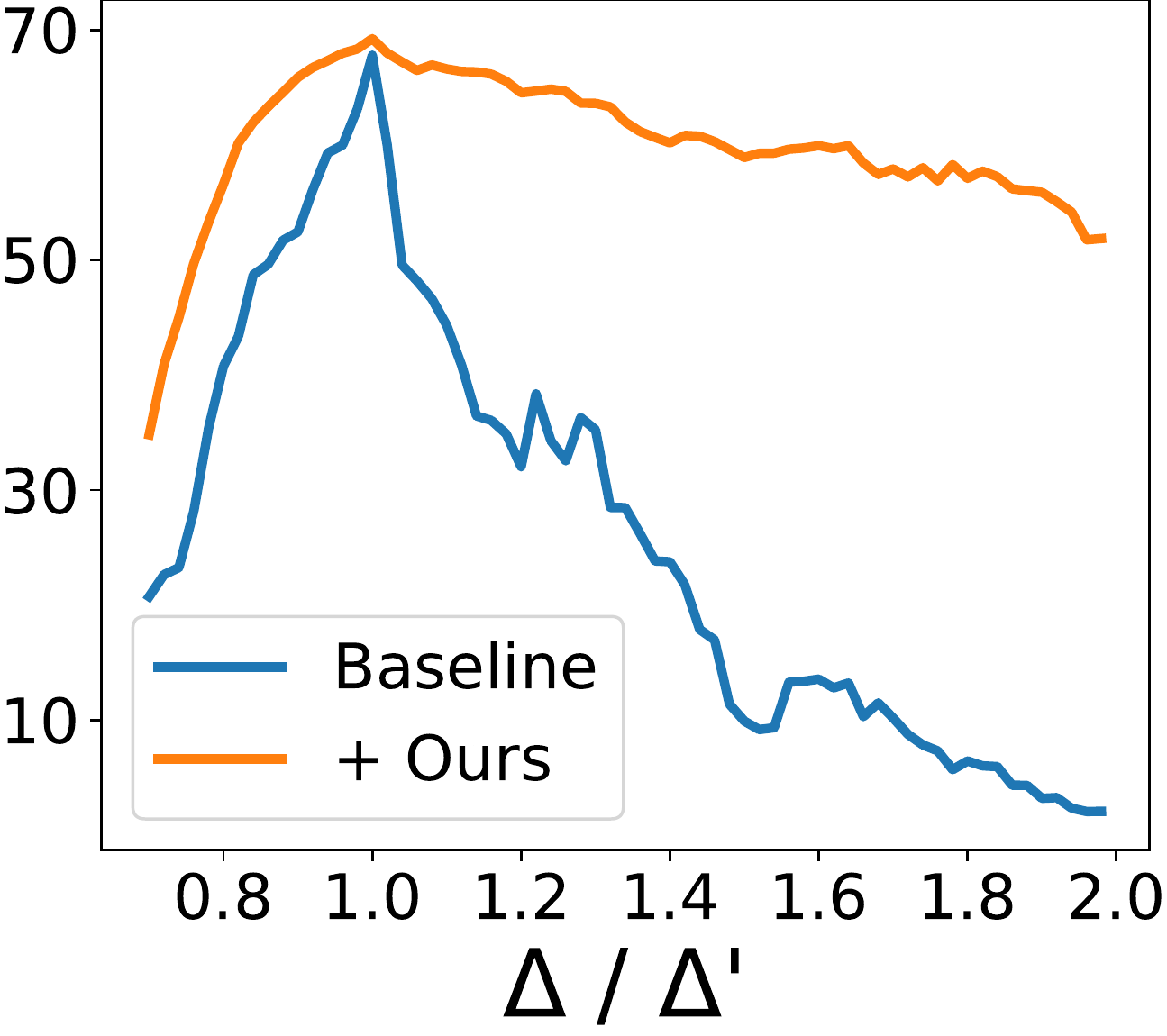}
  \captionsetup{justification=centering}
    \caption{ResNet-18\\PTQ 6b, ImageNet}
\end{subfigure}
\begin{subfigure}{.24\textwidth}
\centering
  \includegraphics[trim= 0mm 5mm 0mm 0mm, width=1.0\linewidth]{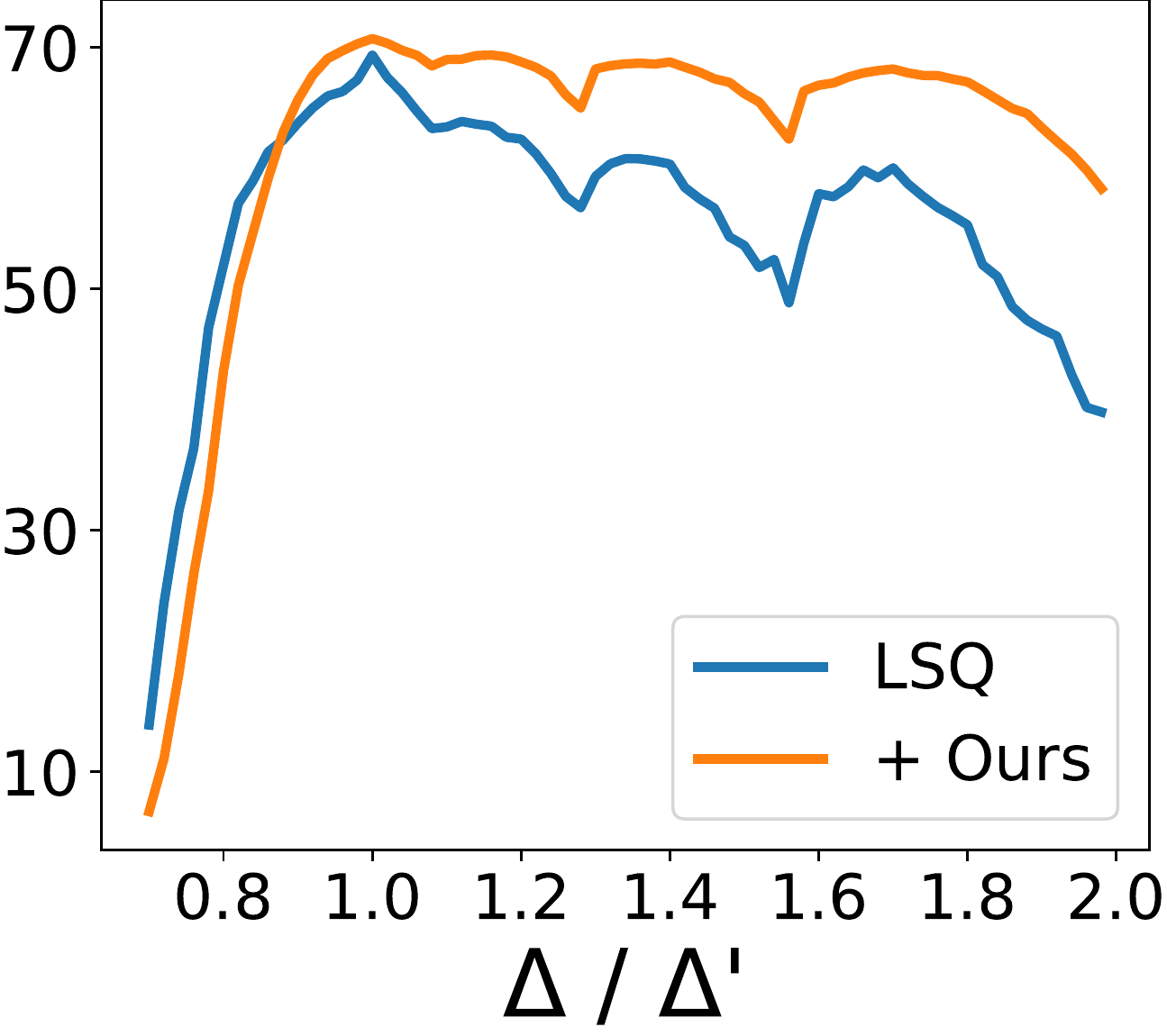}
  \captionsetup{justification=centering}
\caption{ResNet-18\\QAT, ImageNet}
\end{subfigure}
\begin{subfigure}{.24\textwidth}
\centering
  \includegraphics[trim= 0mm 5mm 0mm 0mm, width=1.0\linewidth]{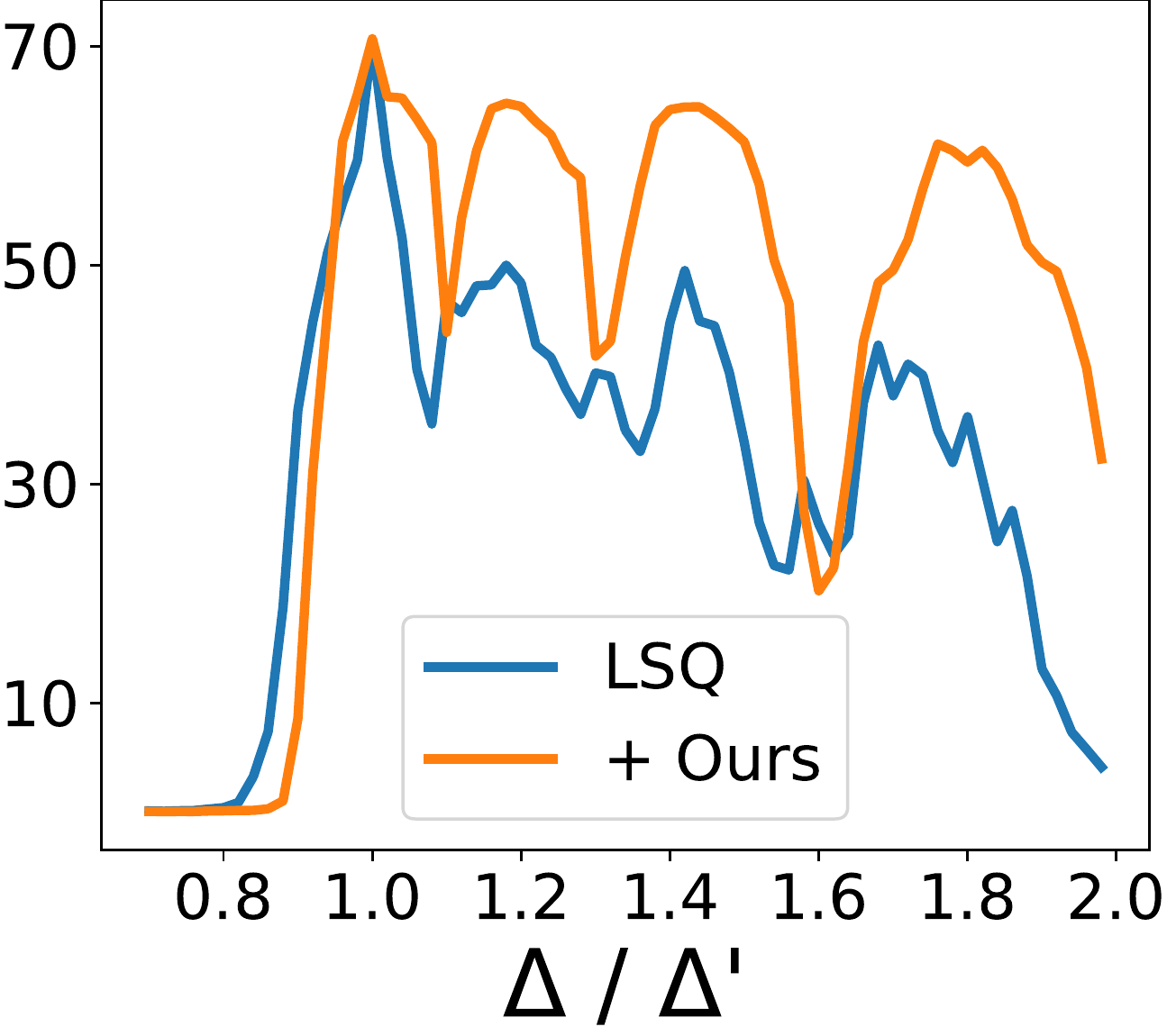}
  \captionsetup{justification=centering}
      \caption{MobileNet-V2\\QAT, ImageNet}
\end{subfigure}

\caption{Robustness of quantized network when we change step size of quantization operator for weight. The networks are optimized for the step size $\Delta'$, and the accuracy is measured with the scaled step size $\Delta$. All networks are quantized into 4-bit (except for (f) into 6-bit) with PTQ~\cite{AdaQuant} and QAT~\cite{Esser2020lsq}, including activation and weight. Additional results are included in the supplementary material. }
\label{fig:exp_rob1}
\end{figure}


To validate the effect of the proposed methods, we measure the accuracy changes depending on the step size as shown in \cref{fig:exp_rob1}. In all cases, including PTQ and QAT, the proposed methods maintain the quality of output in the various step sizes. In the case of MobileNet-V2 for the ImageNet dataset, ours maintains 25.72 \% higher accuracy compared to the baseline when the step size is changed by 8 \%. In addition, ours shows comparable or better results to the previous best method, KURE, for the optimized network (i.e., MobileNet-V2). As indicated in the previous studies~\cite{shkolnik2020robust,jacob2018integer}, the degree of freedom for the quantization step size could be restricted depending on the hardware implementation. For instance, some hardware supports step sizes having predefined values or limited resolution. In such a case, the robustness of the quantization step size is essential to maintain output quality after PTQ or QAT. Because our methods improve the robustness of the step size by a large margin, we expect that our methods could be helpful for the deployment of quantized networks in practice. 


\subsection{Robustness of Bit Precision for QAT}

\begin{figure}[t]
\begin{subfigure}{.26\textwidth}
  \centering
  \includegraphics[trim= 0mm 5mm 0mm 2mm, width=1.0\linewidth]{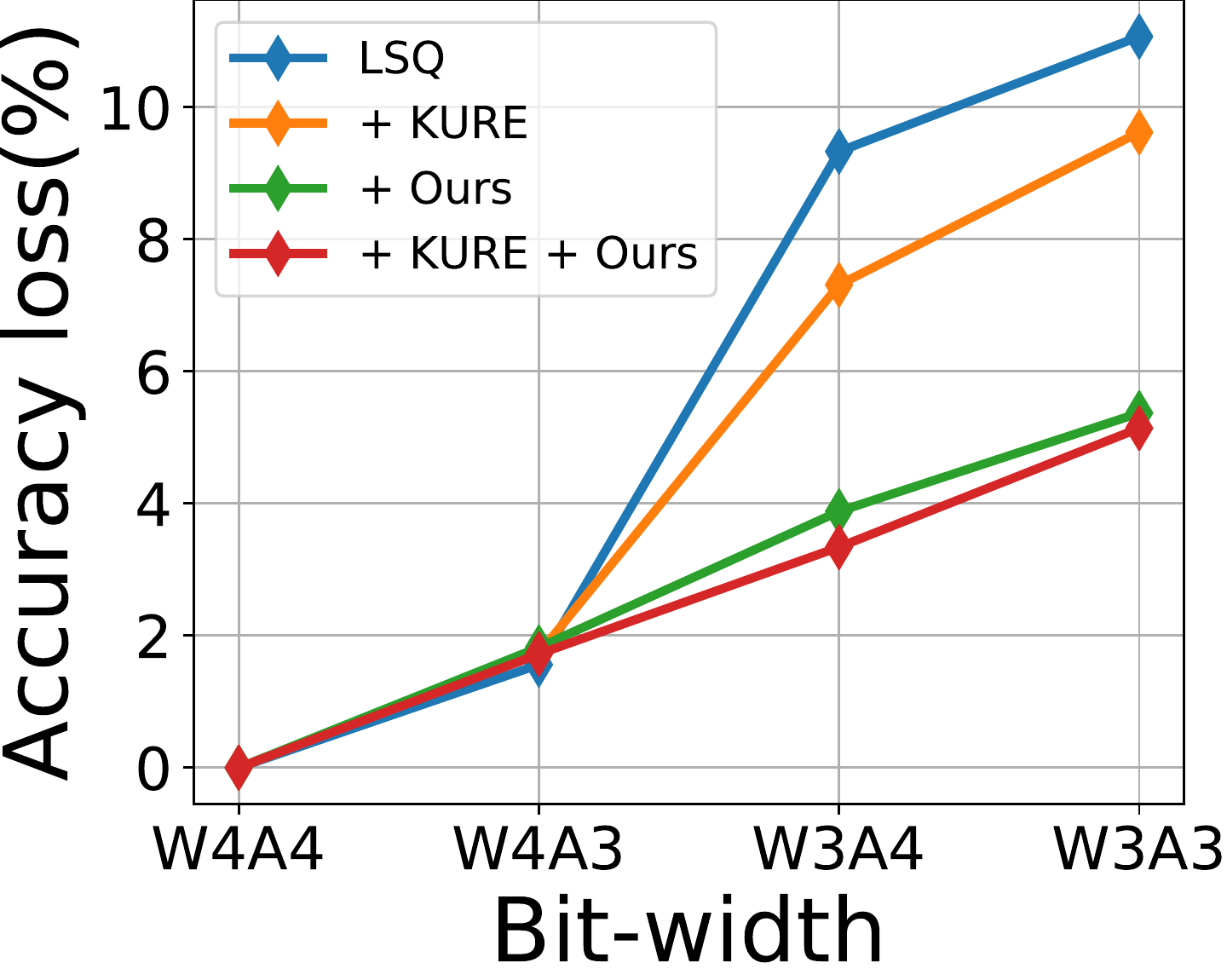}
    \captionsetup{justification=centering}
  \caption{MobileNet-V2\\CIFAR-100, (4/4)  }
  \label{fig:mv2_cifar_44}
\end{subfigure}
\begin{subfigure}{.24\textwidth}
  \centering
  \includegraphics[trim= 0mm 5mm 0mm 5mm, width=1.0\linewidth]{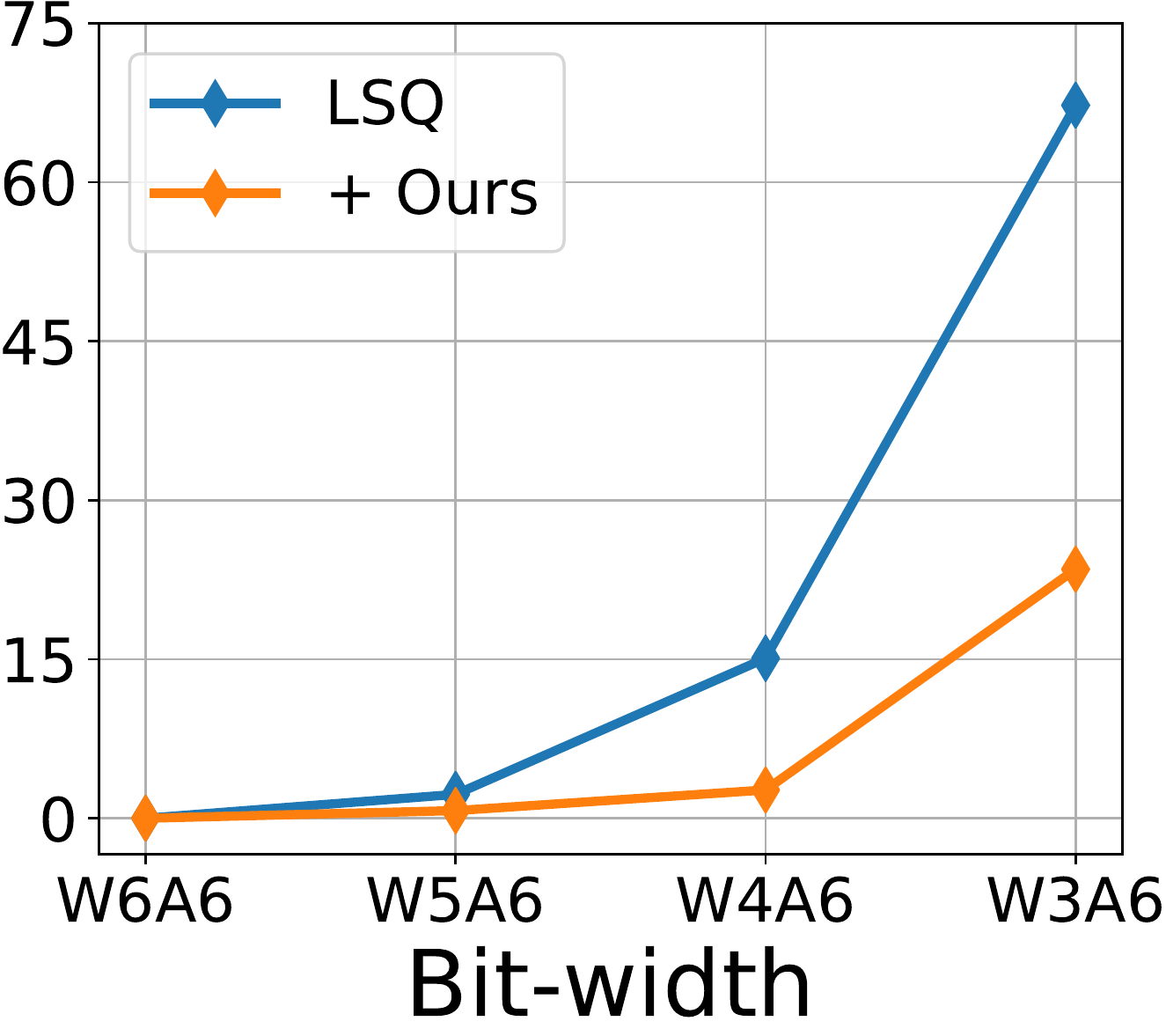}
  \captionsetup{justification=centering}
  \caption{ResNet-18\\ImageNet, (6/6)}
  \label{fig:r18_66}
\end{subfigure}
\begin{subfigure}{.24\textwidth}
  \centering
  \includegraphics[trim= 0mm 5mm 0mm 5mm, width=1.0\linewidth]{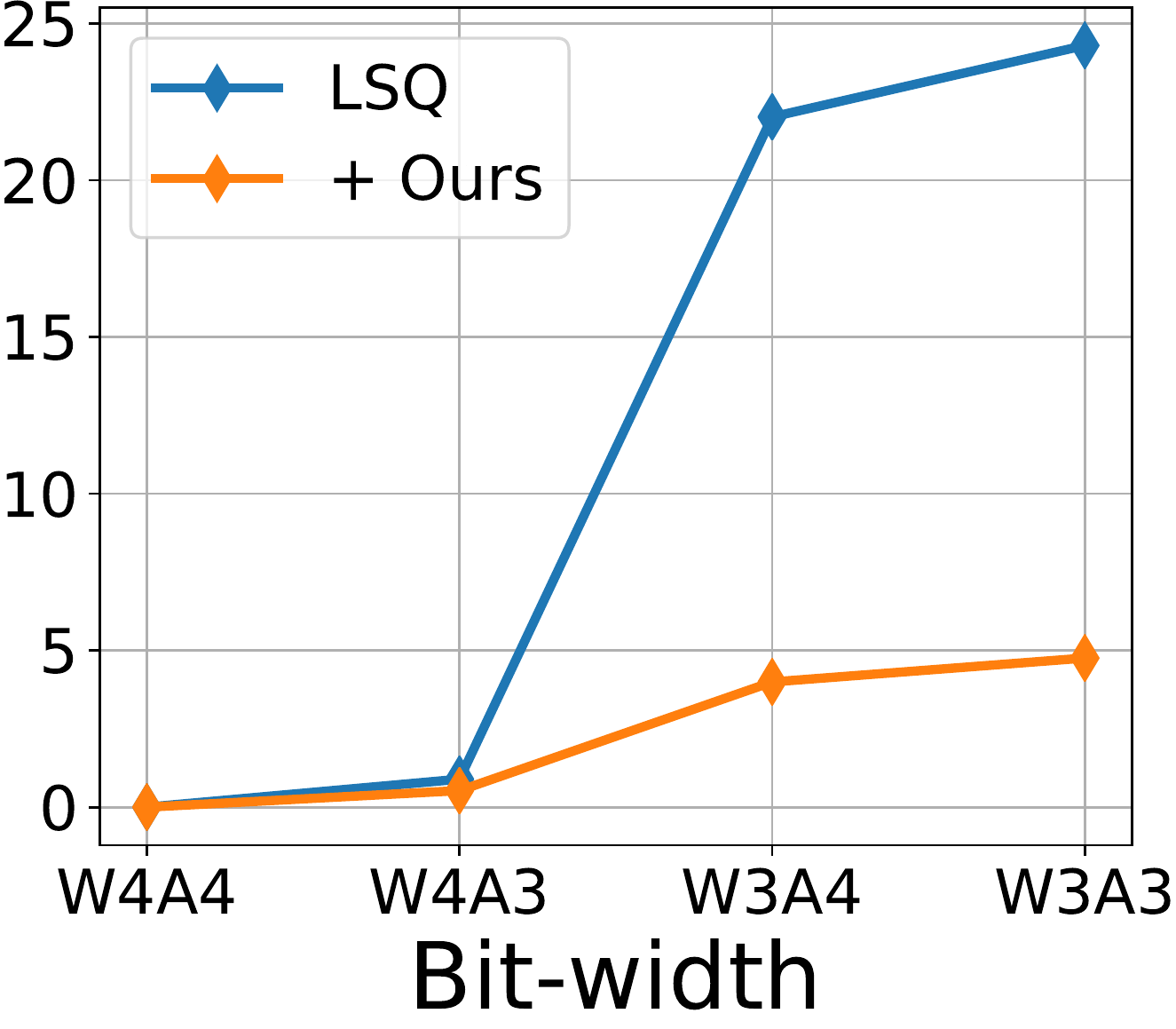}
\captionsetup{justification=centering}
  \caption{ResNet-18\\ImageNet, (4/4)}
   \label{fig:r18_44}
\end{subfigure}
\begin{subfigure}{.24\textwidth}
  \centering
  \includegraphics[trim= 0mm 5mm 0mm 5mm, width=1.0\linewidth]{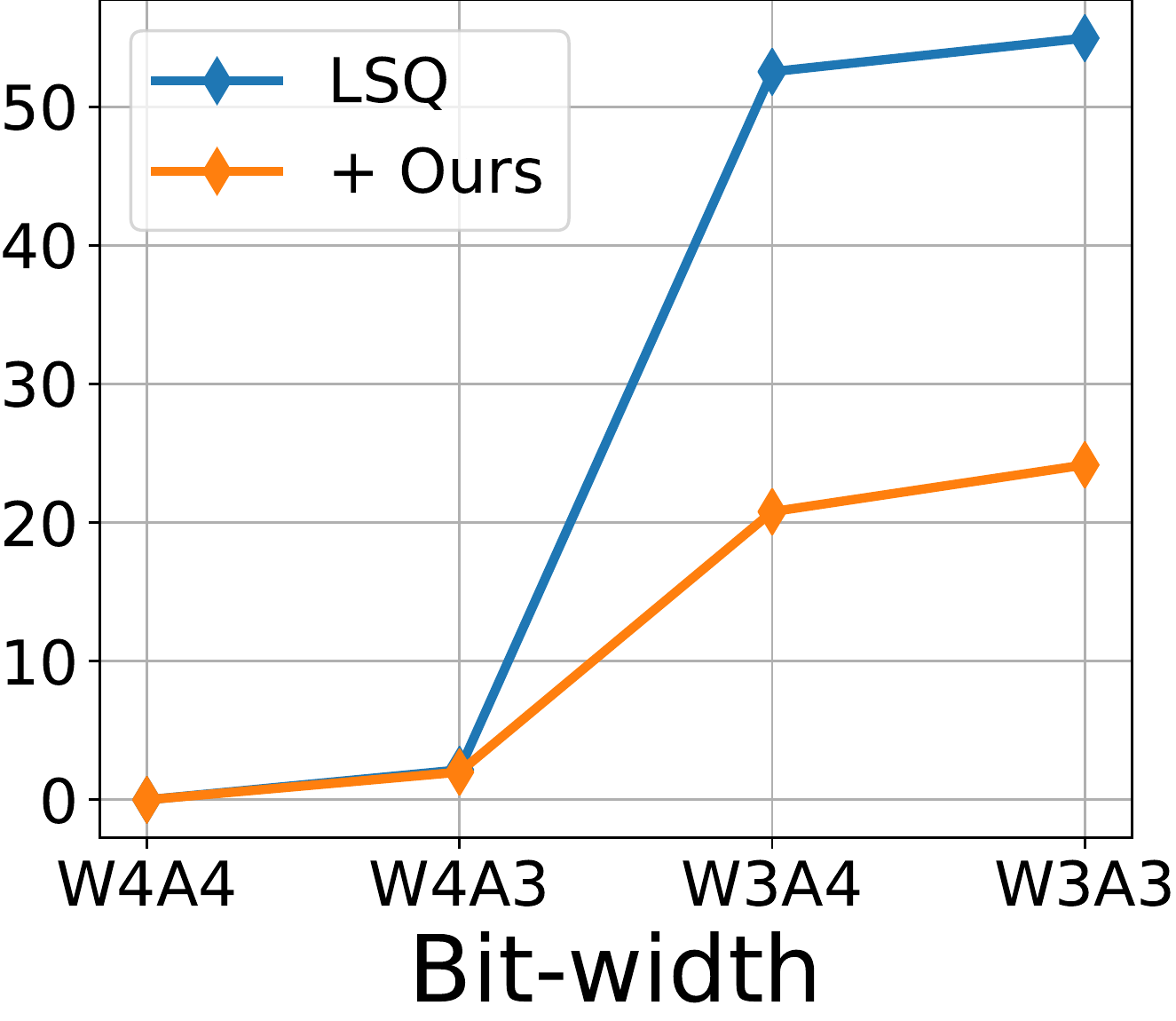}
\captionsetup{justification=centering}
  \caption{MobileNet-V2\\ImageNet, (4/4)}
  \label{fig:mv2_44}
\end{subfigure}
\caption{Robustness of QAT model with and without proposed methods. (W/A) represents the initial bit-width of activation and weight, WXAY indicates the changed bit-width. The accuracy is measured without additional fine-tuning.}
\label{fig:exp_rob2}
\end{figure}


\begin{table}[b]
\caption{Effect of our methods for top-1 accuracy of ResNet-18 and MobileNet-V2 quantized by LSQ~\cite{Esser2020lsq} on the ImageNet dataset (90 epochs of fine-tuning).}
\centering
\resizebox{0.6\textwidth}{!}{

\small
\begin{tabular}{cccccc}
\hline
\multicolumn{1}{c|}{Model}              & \multicolumn{1}{c|}{FP}                     & \multicolumn{1}{c|}{Fine-tuning} & \multicolumn{1}{c|}{4/4}   & \multicolumn{1}{c|}{3/3}   & 2/2   \\ \hline
\multicolumn{1}{c|}{\multirow{2}{*}{ResNet-18}} & \multicolumn{1}{c|}{\multirow{2}{*}{70.542}} & \multicolumn{1}{c|}{LSQ}         & \multicolumn{1}{c|}{69.39} & \multicolumn{1}{c|}{68.80} & 66.26 \\
\multicolumn{1}{c|}{}                          & \multicolumn{1}{c|}{}                       & \multicolumn{1}{c|}{LSQ + Ours}  & \multicolumn{1}{c|}{70.74} & \multicolumn{1}{c|}{69.73} & 66.58 \\ \hline

\multicolumn{1}{c|}{\multirow{2}{*}{MobileNet-V2}} & \multicolumn{1}{c|}{\multirow{2}{*}{72.24}} & \multicolumn{1}{c|}{LSQ}         & \multicolumn{1}{c|}{70.46} & \multicolumn{1}{c|}{67.51} & 44.87 \\
\multicolumn{1}{c|}{}                          & \multicolumn{1}{c|}{}                       & \multicolumn{1}{c|}{LSQ + Ours}  & \multicolumn{1}{c|}{71.16} & \multicolumn{1}{c|}{66.93} & 43.41  \\ \hline

\hline
\end{tabular}
}
\label{tab:qat_table}
\end{table}

Because most of the existing QAT methods specialize the weight fine-tuning for the specific configuration, the accuracy of the quantized network in different bit-widths is reduced significantly. Meanwhile, ours enhance the robustness of the quantized network, allowing stable accuracy in different bit-widths without fine-tuning. \cref{fig:exp_rob2}. shows the accuracy degradation depending on the operation bit-widths other than the one we trained via QAT. With the proposed methods, one can train a generic model that can produce reliable output in multiple bit-widths with existing QAT algorithms.

Table \ref{tab:qat_table} shows the accuracy of QAT for ResNet-18/MobileNet-V2 in different bit-widths with/without the proposed methods. As shown in the table, the proposed fine-tuning scheme does not reduce the accuracy of the quantized network in all cases of ResNet-18 and 4-bit MobileNet-V2. In the case of 3-/2-bit MobileNet-V2, we speculated that the accuracy is slightly degraded due to the overlapped effect of strong regularization and the limited degree of freedom. When applying quantization with 4-bit or higher precision, we can enjoy the benefit of robustness based on the proposed methods without losing accuracy.

\section{Discussion}\label{sec:limit}
Our proposed methods are advantageous in minimizing quantization error after PTQ and QAT. However, one remaining important topic that we could not address in this paper is the robustness of the activation. In this paper, we rely on the PTQ/QAT algorithms for activation quantization. Unlike weight, activation is input-dependent, and the distribution is diverse depending on the behavior of nonlinear functions. This is a challenging problem and left as future work.


\section{Conclusion}\label{sec:con}
Enhancing the robustness of neural networks for quantization maximizes the benefit we can get from the low-precision operations. In this study, we reported three important motivations for minimizing the accuracy degradation after quantization: reduction of error propagation, range clamping for error minimization, and inherited robustness against quantization. Based on these insights, we proposed two novel ideas, symmetry regularization (SymReg) and saturating nonlinearity (SatNL). Our extensive experiments verified the advantages of the proposed methods, which significantly reduce the quantization error of diverse QAT and PTQ algorithms. Enhancing the robustness of quantization is achievable with negligible extra cost, but it enables us to exploit the benefit of low-precision computation with minimal accuracy degradation. We expect that the robustness of networks will minimize the deployment overhead for energy-efficient NPUs, thereby positively affecting the environment.

\subsubsection*{Acknowledgements.} This work was supported by IITP grant funded by the Korea government (MSIT, No.2019-0-01906, No.2021-0-00105, and No.2021-0-00310), SK Hynix Inc. and Google Asia Pacific.

%
%
\bibliographystyle{splncs04}

\end{document}


\pagestyle{headings}
\mainmatter
\def\ECCVSubNumber{2337}  

\title{Symmetry Regularization and Saturating Nonlinearity for Robust Quantization \\ Supplementary Materials} 

\titlerunning{ECCV-22 submission ID \ECCVSubNumber} 
\authorrunning{ECCV-22 submission ID \ECCVSubNumber} 
\author{Anonymous ECCV submission}
\institute{Paper ID \ECCVSubNumber}

\maketitle

\section{Error Propagation Comparison}
\begin{figure}
  \centering
   \includegraphics[width=0.8\linewidth]{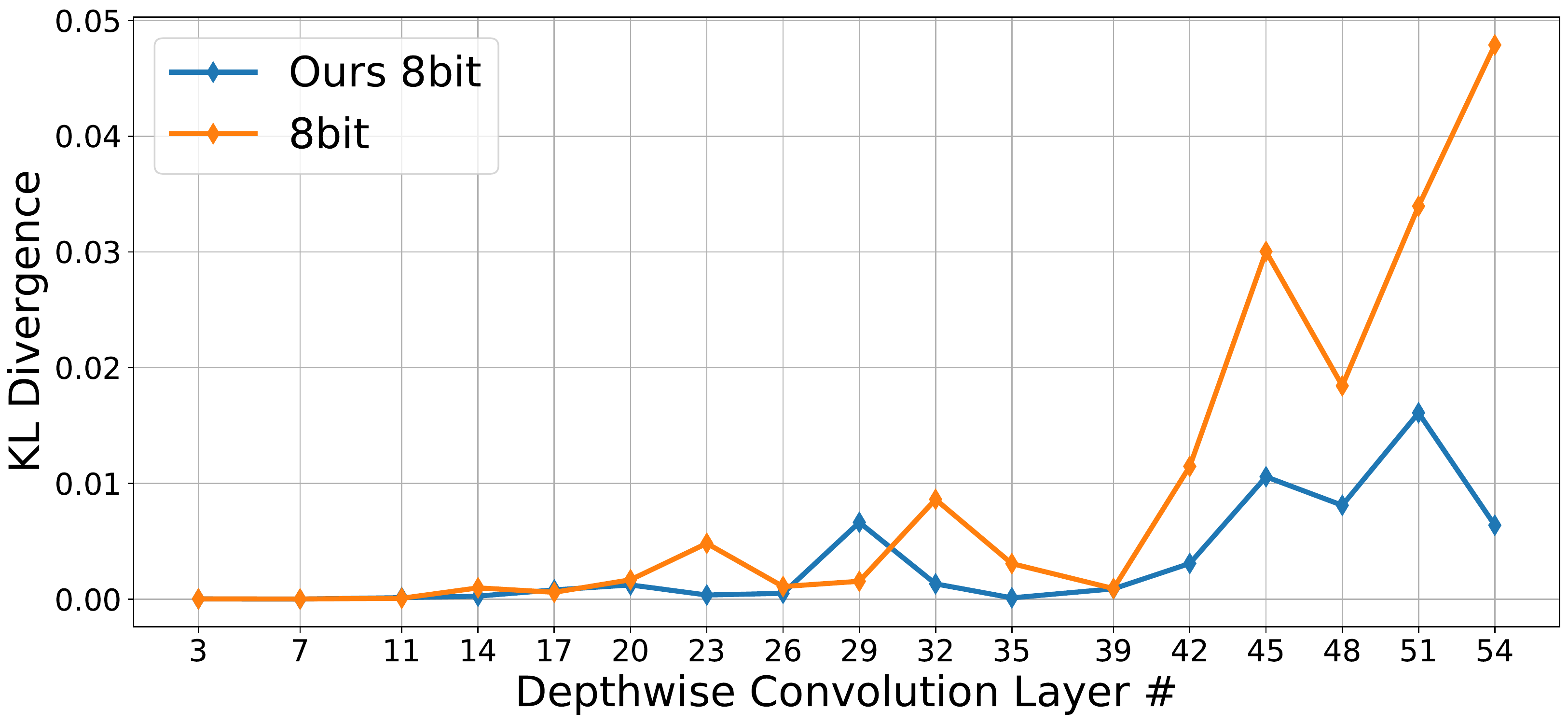}
   \caption{KL Divergence of depthwise convolution output between the baseline MobileNet-V2 and the model with ours (e.g., SymReg, SatNL, and ASAM) on the ImageNet dataset. The weights of both networks are quantized into 8-bit through PTQ (ACIQ).}
   \label{fig:KL_Propa}
\end{figure}

\cref{fig:KL_Propa} shows the layer-wise KL divergence of output activation before and after the 8-bit weight quantization. As shown in the figure, the KL divergence of the baseline network becomes larger in the last depthwise convolution layer, while the divergence of the proposed network has a much smaller difference. Because the proposed methods minimize the quantization error through SatNL, the layer-wise error is smaller than the original network. In addition, SymReg mitigates the error propagation, which prevents the accumulation of quantization errors over multiple layers. As a result, the output activation could maintain the consistent features, and we could enjoy the benefit of low-precision computation with minimal accuracy degradation.

\begin{figure}[t]
  \centering
   \includegraphics[width=0.8\linewidth]{./figure/all_2 (1).pdf}
   \caption{Weight distribution of convolution kernels in the 15th convolution layer of MobileNet-V2 at CIFAR-100. Left : Baseline, Middle : SatNL, Right : SatNL+SymReg.}
   \label{fig:weightdistribution}
\end{figure}

\section{Weight Distribution Visualization}
\cref{fig:weightdistribution} shows the effect of the proposed methods by visualizing the histograms of weights in the 15th convolution layer of MobileNet-v2 at CIFAR-100. As shown in the left figure, the original weight has an irregular distribution with a few large values. Due to these infrequent values, the quantization error is increased after the quantization. When we train the network with SatNL, the distributions are concentrated within the narrowed range, as shown in the middle figure. As a result, the statistics difference before and after the quantization could be minimized. After applying SymReg in addition to SatNL, the distribution now becomes symmetric, and thereby the biased quantization error is forced to zero regardless of quantization algorithms. While the proposed methods reduce the degree of freedom of weight, introducing minor accuracy degradation, the robustness of the network against quantization could be enhanced significantly.

\section{Training Configurations}
\begin{table*}
\centering
\begin{tabular}{@{}cclcc|cc|c@{}}
\toprule
\multicolumn{1}{l}{}                              &                               &       &     &                  & \multicolumn{2}{c|}{Cosine annealing with warmup} & ASAM   \\ \midrule
\multicolumn{2}{c|}{configuration}                                                & epoch & lr  & weight decay     & warmup len           & $\eta_{min}$               & $\rho$ \\ \midrule
\multicolumn{1}{c|}{ResNet18}                     & \multicolumn{1}{c|}{ImageNet} & 150   & 0.4 & $1\times10^{-5}$ & 5                    & $1\times10^{-2}$           & 1      \\ \midrule
\multicolumn{1}{c|}{\multirow{2}{*}{MobileNetv2}} & \multicolumn{1}{c|}{Cifar100} & 120   & 0.4 & $5\times10^{-5}$ & 5                    & $1\times10^{-2}$           & 1      \\
\multicolumn{1}{c|}{}                             & \multicolumn{1}{c|}{ImageNet} & 150   & 0.4 & $1\times10^{-5}$ & 5                    & $1\times10^{-2}$           & 1      \\ \midrule
\multicolumn{1}{c|}{MobileNetv3}                  & \multicolumn{1}{c|}{ImageNet} & 240   & 0.4 & $1\times10^{-5}$ & 5                    & $1\times10^{-2}$           & 0.2    \\ \bottomrule
\end{tabular}
\caption{Hyper-parameters to train the networks}
\label{tab:training_config}
\end{table*}
In this work, we need to train the target models from scratch for PTQ experiments. \cref{tab:training_config} shows the hyper-parameters we use to train the models. We use the well-known SGD with a momentum algorithm and the exponential moving average of parameters. In the case of QAT, we apply 90 epochs of fine-tuning with 1/10 lower learning rates than used in the pre-training stage. The rest of the hyper-parameters are set identical to the initial full-precision pre-training stage. 

\section{Reproducibility}
We attach the source code of experiments in addition to this supplementary article. After the review period, the entire source code will be re-factored and available in the author's public Github repository.

\section{Other Non-linearities for SatNL}
\begin{figure}
  \centering
  \includegraphics[width=0.6\linewidth]{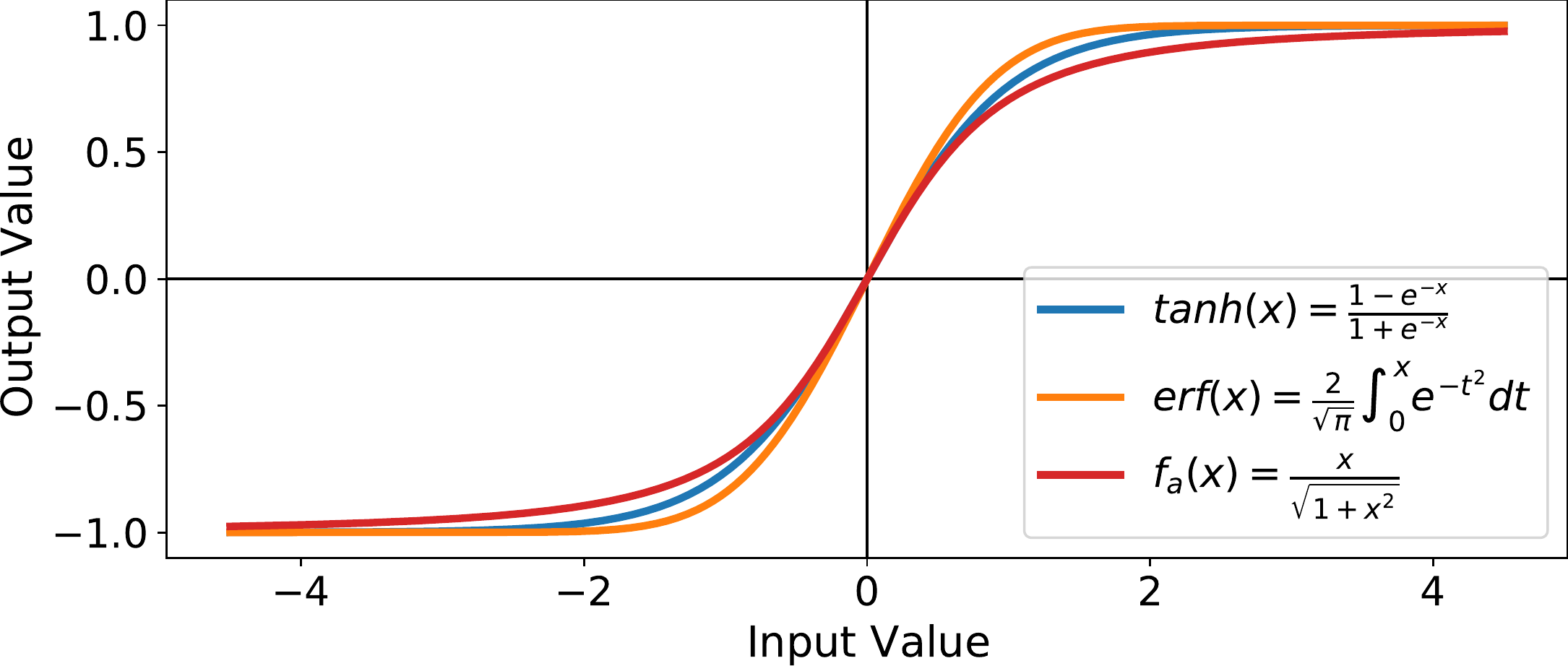}
   \caption{Various saturating non-linearity functions.}
  
   \label{fig:nl_function}
\end{figure}

As mentioned in section 4.2, SatNL requires three properties; 1. odd function, 2. bounded range, and 3. decreased slope. We reproduce the experiments of Table 1 in the main paper by replacing $tanh$ with two other functions satisfying the properties (\cref{fig:nl_function}). We observe that the accuracy difference is negligible, proving that the choice of the SatNL function has no notable impact on the accuracy.

\section{ImageNet Experiments with KURE}
\cref{tab:mainresult} extends Table 1 in the main paper by including the comparison to the previous state-of-the-art method, KURE\cite{shkolnik2020robust}. As shown in the table, the joint regularization of ours and KURE shows the highest accuracy in extreme low-precision PTQ (3-bit) compared to other methods. However, because KURE is a strong regularization, it introduces slight accuracy degradation in the full-precision pre-training stage. When applying PTQ with 4-bit or higher precision, ours without KURE shows higher accuracy in every case. Our novel ideas push the boundaries of achievable accuracy when one quantizes the network with PTQ in 4-bit or higher.

\definecolor{LightCyan}{rgb}{0.95,0.95,0.95}

\begin{table*}[ht!]
\centering
\resizebox{\hsize}{!}{
\begin{tabular}{ccccccccccc}
\hline
\multirow{2}{*}{Model}                            & \multirow{2}{*}{PTQ}                           & \multirow{2}{*}{Method}       & \multicolumn{8}{c}{Weight/activation bit-width configuration}                             \\ \cline{4-11} 
                                                  &                                                &                               & FP    & 4/FP   & 3/FP   & 2/FP                        & 6/6    & 5/5    & 4/4    & 3/3    \\

\hline

\multicolumn{1}{c|}{\multirow{12}{*}{ResNet-18}}   & \multicolumn{1}{c|}{\multirow{4}{*}{ACIQ}}     & \multicolumn{1}{c|}{Baseline} & 70.54 & 47.44  & -      & \multicolumn{1}{c|}{-}      & 68.70   & 64.87 & 38.46 & -      \\
\multicolumn{1}{c|}{}                             & \multicolumn{1}{c|}{}                          & \multicolumn{1}{c|}{Ours}     & 70.92 & 69.22 & 49.06 & \multicolumn{1}{c|}{-}      & 70.02  & 68.99 & 66.65 & 42.95 \\
\multicolumn{1}{c|}{}                             & \multicolumn{1}{c|}{}                          &   \multicolumn{1}{c|}{KURE}     & \rowcolor{LightCyan} 69.39 &  66.69  & 44.19 & \multicolumn{1}{c|}{-}      & 68.01 & 66.84 & 61.80 & 26.58      \\
\multicolumn{1}{c|}{}                             & \multicolumn{1}{c|}{}                          & \multicolumn{1}{c|}{Ours+KURE}  & \rowcolor{LightCyan} 70.33 &  69.85  & 67.59 & \multicolumn{1}{c|}{-}      & 69.23 & 68.65 & 67.11 & 61.07      \\

\cline{2-11} 
\multicolumn{1}{c|}{}                             & \multicolumn{1}{c|}{\multirow{4}{*}{AdaQuant}} & \multicolumn{1}{c|}{Baseline} & 70.54 & 69.29 & 66.18 & \multicolumn{1}{c|}{3.23}  & 70.17 & 69.55 & 67.67 & 57.57 \\
\multicolumn{1}{c|}{}                             & \multicolumn{1}{c|}{}                          & \multicolumn{1}{c|}{Ours}     & 70.92 & 70.36 & 68.84 & \multicolumn{1}{c|}{48.39} & 70.75 & 70.37  & 69.35 & 64.04 \\
\multicolumn{1}{c|}{}                             & \multicolumn{1}{c|}{}                          & \multicolumn{1}{c|}{KURE}     & \rowcolor{LightCyan} 69.39 &  69.09  & 68.32 & \multicolumn{1}{c|}{62.21}      & 69.23 & 68.77 & 67.77 & 64.22      \\
\multicolumn{1}{c|}{}                             & \multicolumn{1}{c|}{}                          & \multicolumn{1}{c|}{Ours+KURE}     & \rowcolor{LightCyan} 70.33 &  69.96  & 69.21 & \multicolumn{1}{c|}{63.16}      & 70.11 & 69.77 & 69.14 & 66.06      \\

\cline{2-11} 
\multicolumn{1}{c|}{}                             & \multicolumn{1}{c|}{\multirow{4}{*}{QDrop}} & \multicolumn{1}{c|}{Baseline} &  70.54 & 70.15 & 69.39 & \multicolumn{1}{c|}{66.40}  & 70.27 & 69.93 & 68.91 & 65.75 \\
\multicolumn{1}{c|}{}                             & \multicolumn{1}{c|}{}                          & \multicolumn{1}{c|}{Ours}     & 70.92 & 70.69 & 70.062 & \multicolumn{1}{c|}{66.98} & 70.81 & 70.57 & 69.93 & 67.45  \\
\multicolumn{1}{c|}{}                             & \multicolumn{1}{c|}{}                          & \multicolumn{1}{c|}{KURE}     & \rowcolor{LightCyan} 69.39 & 69.35  & 69.14 & \multicolumn{1}{c|}{67.60}      & 69.25  & 69.13 & 68.30 & 66.10       \\
\multicolumn{1}{c|}{}                             & \multicolumn{1}{c|}{}                          & \multicolumn{1}{c|}{Ours+KURE}     & \rowcolor{LightCyan} 70.33 & 70.18  & 69.86 & \multicolumn{1}{c|}{67.77}      & 70.13 & 69.86 & 69.38 & 67.47         \\

\hline

\multicolumn{1}{c|}{\multirow{12}{*}{MobileNet-V2}} & \multicolumn{1}{c|}{\multirow{4}{*}{ACIQ}}    & \multicolumn{1}{c|}{Baseline} & 72.22 & 28.68 & -      & \multicolumn{1}{c|}{-}      & 69.30 & 64.20 & 18.15      & -      \\
\multicolumn{1}{c|}{}                             & \multicolumn{1}{c|}{}                          & \multicolumn{1}{c|}{Ours}     & 72.87 & 70.07 & 40.79 & \multicolumn{1}{c|}{-}      & 71.07 & 68.66 & 58.30 & 6.25      \\
\multicolumn{1}{c|}{}                             & \multicolumn{1}{c|}{}                          & \multicolumn{1}{c|}{KURE}     & \rowcolor{LightCyan} 72.07 &  54.34  & 6.43 & \multicolumn{1}{c|}{-}      & 69.77 & 64.37 & 39.31 & 2.14      \\
\multicolumn{1}{c|}{}                             & \multicolumn{1}{c|}{}                          & \multicolumn{1}{c|}{Ours+KURE}     & \rowcolor{LightCyan}72.48 &  70.30  & 42.24 & \multicolumn{1}{c|}{-}      & 70.68 & 68.31 & 61.51 & 13.87      \\

\cline{2-11}
\multicolumn{1}{c|}{}                             & \multicolumn{1}{c|}{\multirow{4}{*}{AdaQuant}} & \multicolumn{1}{c|}{Baseline} & 72.22 & 70.67 & 59.80 & \multicolumn{1}{c|}{-}      & 71.52 & 70.72 & 63.70 & -      \\
\multicolumn{1}{c|}{}                             & \multicolumn{1}{c|}{}                          & \multicolumn{1}{c|}{Ours}     & 72.87 & 72.23 & 69.03  & \multicolumn{1}{c|}{-}      & 72.27 & 71.76 & 68.91 & 18.36 \\
\multicolumn{1}{c|}{}                             & \multicolumn{1}{c|}{}                          & \multicolumn{1}{c|}{KURE}     & \rowcolor{LightCyan}72.07 & 71.51   & 68.71 & \multicolumn{1}{c|}{3.58}      & 71.62 & 70.71 & 66.25 & 4.61      \\
\multicolumn{1}{c|}{}                             & \multicolumn{1}{c|}{}                          & \multicolumn{1}{c|}{Ours+KURE}     & \rowcolor{LightCyan} 72.48 &  71.93 & 70.17 & \multicolumn{1}{c|}{10.16}      & 71.90 & 71.26 & 68.84 & 29.27     \\
\cline{2-11} 
\multicolumn{1}{c|}{}                             & \multicolumn{1}{c|}{\multirow{4}{*}{QDrop}} & \multicolumn{1}{c|}{Baseline} & 72.22 & 71.41 & 68.32 & \multicolumn{1}{c|}{48.68}  & 71.57 & 70.64 & 67.08 & 50.79 \\
\multicolumn{1}{c|}{}                             & \multicolumn{1}{c|}{}                          & \multicolumn{1}{c|}{Ours}     & 72.87 & 72.44 & 71.18 & \multicolumn{1}{c|}{61.68} & 72.61 & 72.05 & 69.87 & 62.55 \\
\multicolumn{1}{c|}{}                             & \multicolumn{1}{c|}{}                          & \multicolumn{1}{c|}{KURE}     & \rowcolor{LightCyan} 72.07 & 71.75 & 70.55  & \multicolumn{1}{c|}{59.51}      & 71.76 & 70.91 & 68.23 & 56.20      \\
\multicolumn{1}{c|}{}                             & \multicolumn{1}{c|}{}                          & \multicolumn{1}{c|}{Ours+KURE}     & \rowcolor{LightCyan} 72.48 & 72.13  & 71.25 & \multicolumn{1}{c|}{63.78}      & 72.15 & 71.60 & 70.00 & 63.30     \\

\hline

\multicolumn{1}{c|}{\multirow{8}{*}{MobileNet-V3}} & \multicolumn{1}{c|}{\multirow{4}{*}{ACIQ}}    & \multicolumn{1}{c|}{Baseline} & 74.52 & 29.65      & -      & \multicolumn{1}{c|}{-}      & -      & -      & -      & -      \\
\multicolumn{1}{c|}{}                             & \multicolumn{1}{c|}{}                          & \multicolumn{1}{c|}{Ours}     & 74.43 & 61.95      & 1.04      & \multicolumn{1}{c|}{-}      & -      & -      & -      & -      \\
\multicolumn{1}{c|}{}                             & \multicolumn{1}{c|}{}                          & \multicolumn{1}{c|}{KURE}     & \rowcolor{LightCyan} 73.81 &  55.15  & 3.09 & \multicolumn{1}{c|}{-}      & - & - & - & -      \\
\multicolumn{1}{c|}{}                             & \multicolumn{1}{c|}{}                          & \multicolumn{1}{c|}{Ours+KURE}     & \rowcolor{LightCyan}73.84 &  66.21  & 5.44 & \multicolumn{1}{c|}{-}      & - & - & - & -      \\

\cline{2-11} 
\multicolumn{1}{c|}{}                             & \multicolumn{1}{c|}{\multirow{4}{*}{AdaQuant}} & \multicolumn{1}{c|}{Baseline} & 74.52 & 72.92 & 64.17 & \multicolumn{1}{c|}{-}      & 72.73 & 68.95 & 43.88 & -      \\
\multicolumn{1}{c|}{}                             & \multicolumn{1}{c|}{}                          & \multicolumn{1}{c|}{Ours}     & 74.43 & 73.51 & 70.50 & \multicolumn{1}{c|}{2.87}      & 72.69 & 71.02 & 62.73 & -      \\
\multicolumn{1}{c|}{}                             & \multicolumn{1}{c|}{}                          & \multicolumn{1}{c|}{KURE}     & \rowcolor{LightCyan}73.81 &  73.11  & 70.46 & \multicolumn{1}{c|}{7.44}      & 72.63 & 70.63 & 59.25 & -      \\
\multicolumn{1}{c|}{}                             & \multicolumn{1}{c|}{}                          & \multicolumn{1}{c|}{Ours+KURE}     & \rowcolor{LightCyan}73.84 &  73.25  & 70.91 & \multicolumn{1}{c|}{20.05}      & 72.29 & 70.32 & 62.35 & 1.84      \\

\hline
\end{tabular}
}
\caption{Results of applying PTQ to baseline and network with proposed ideas including KURE on ImageNet dataset. The values in the table represent the top-1 accuracy. The dashed cells represent the points where the PTQ fails to converge, having lower than 1~\% of accuracy.}
\label{tab:mainresult}
\end{table*}

\section{Additional Experiments on Non-linear PTQ Algorithms}
In order to show the outstanding benefit of the proposed methods for the non-linear quantization algorithms, we conducted extensive studies based on non-linear PTQ algorithms (i.e., logarithm-based quantization and K-means clustering-based quantization). As shown in \cref{fig:exp_robust}, our proposed methods increase the robustness by a large margin in both methods, allowing minimal accuracy degradation in low-precision. This result verifies that the proposed methods are also applicable for non-linear quantization. Compared to the previous best, KURE, ours gives comparable or slightly better robustness in the optimized networks, i.e., MobileNet-V2/V3.  According to our observation, SatNL is highly beneficial to stabilize the non-linear PTQ process because the statistical difference before and after the quantization could be minimized.

\begin{figure}
\begin{subfigure}{.34\textwidth}
    \centering
  \includegraphics[width=1.0\linewidth]{figure/Resnet_imagenet_codebookquant_sym (8).pdf}
  \captionsetup{justification=centering}
  \caption{ResNet-18\\K-means, ImageNet}
\end{subfigure}
\begin{subfigure}{.31\textwidth}
    \centering
  \includegraphics[width=1.0\linewidth]{figure/mv2_imagenet_codebookquant_sym (5).pdf}
  \captionsetup{justification=centering}
  \caption{MobileNet-V2\\K-means, ImageNet}
\end{subfigure}
\begin{subfigure}{.31\textwidth}
    \centering
    \includegraphics[width=1.0\linewidth]{figure/mv3_imagenet_codebookquant_sym (5).pdf}
  \captionsetup{justification=centering}
  \caption{MobileNet-V3\\K-means, ImageNet}
\end{subfigure}

\begin{subfigure}{.34\textwidth}
    \centering
  \includegraphics[width=1.0\linewidth]{figure/Resnet_imagenet_logquant (2).pdf}
  \captionsetup{justification=centering}
  \caption{ResNet-18\\Log-Quant, ImageNet}
\end{subfigure}
\begin{subfigure}{.31\textwidth}
    \centering
  \includegraphics[width=1.0\linewidth]{figure/mv2_imagenet_logquant (4).pdf}
  \captionsetup{justification=centering}
  \caption{MobileNet-V2\\Log-Quant, ImageNet}
\end{subfigure}
\begin{subfigure}{.31\textwidth}
    \centering
    \includegraphics[width=1.0\linewidth]{figure/mv3_imagenet_logquant (1).pdf}
  \captionsetup{justification=centering}
  \caption{MobileNet-V3\\Log-Quant, ImageNet}
\end{subfigure}

\caption{Robustness of quantized networks with non-linear quantization algorithms. The weight precision is changed while the activation remains full-precision. }
\label{fig:exp_robust}
\end{figure}

\section{Addtional Results Regarding Robustness for Quantization Step Size}
\cref{fig:exp_rob1a} shows the additional experiments for measuring the robustness of networks for step size changes corresponding to Fig. 7 in the main paper. In all cases, the quantized models with proposed methods maintain the accuracy in various quantization step sizes. Our methods are beneficial for robust quantization even for the optimized networks, i.e., MobileNet-V2/V3.

\begin{figure}
\begin{subfigure}{.26\textwidth}
    \centering
  \includegraphics[width=1.0\linewidth]{./figure/ratio_imagenet_mv2_66_ptq (2).pdf}
  \captionsetup{justification=centering}
  \caption{MobileNet-V2\\PTQ 6b, ImageNet}
\end{subfigure}
\begin{subfigure}{.24\textwidth}
    \centering
    \includegraphics[width=1.0\linewidth]{./figure/ratio_imagenet_mv3_66_ptq (3).pdf}
  \captionsetup{justification=centering}
  \caption{MobileNet-V3\\PTQ 6b, ImageNet}
\end{subfigure}
\begin{subfigure}{.24\textwidth}
    \centering
  \includegraphics[width=1.0\linewidth]{figure/ratio_imagenet_resnet18_44_PTQ (7).pdf}
  \captionsetup{justification=centering}
  \caption{ResNet-18\\PTQ 4b, ImageNet}
\end{subfigure}
\begin{subfigure}{.24\textwidth}
    \centering
  \includegraphics[width=1.0\linewidth]{figure/ratio_imagenet_resnet18_66_PTQ (4).pdf}
  \captionsetup{justification=centering}
  \caption{ResNet-18\\PTQ 6b, ImageNet}
\end{subfigure}


\caption{Robustness of quantized network when we change step size of quantization operator for weight. The networks are optimized for the step size $\Delta'$, and the accuracy is measured with the scaled step size $\Delta$. All networks are quantized into the given bitwidth with PTQ~\cite{AdaQuant}, including activation and weight.}
\label{fig:exp_rob1a}
\end{figure}

\section{Explanation of Equations}
\subsection{Equation 4 and 6}
In Equations 4 and 6, we follow the derivation of ACIQ~\cite{banner2018aciq} regarding the expected mean-square-error of linear quantization. When we apply b-bit quantization to the quantization boundaries $[-\alpha, \alpha]$, the quantization interval is equally divided into $2^b$ discrete levels. When the density function is given as $f(x)$, the overall quantization error is expressed as follows:

\begin{equation}
\begin{split}
&\text{Quantization Error} = E[(W - Q(W))^2] \\
&= \overbrace{\int_{-\infty}^{\infty} f(x)\cdot(x-\alpha)^2dx}^{\text{quantization error}} \\
&= \overbrace{\cdot \int_{\alpha}^{\infty} f(x)\cdot(x-\alpha)^2dx + \cdot \int_{-\infty}^{\alpha} f(x)\cdot(x-\alpha)^2dx}^{\text{truncation error}}\\
&+ \overbrace{\sum_{i=0}^{2^M-1}\int_{-\alpha+i\cdot \Delta}^{-\alpha+(i+1)\cdot \Delta} f(x)\cdot (x-q_i)^2dx}^{\text{rounding error}}, 
\end{split}
\end{equation}
where $\Delta = 2\cdot\alpha / 2^b$ and $q_i = -\alpha + (2i+1)\cdot \Delta/2$.

In the previous study~\cite{banner2018aciq}, the rounding error is approximated by a piece-wise linear function based on the slope and the value of the density function at the midpoint of quantization levels, $q_i$. The rounding error of the quantization noise is approximated as follows:
\begin{equation}
\begin{split}
   \overbrace{\sum_{i=0}^{2^M-1}\int_{-\alpha+i\cdot \Delta}^{-\alpha+(i+1)\cdot \Delta} f(x)\cdot (x-q_i)^2dx}^{\text{rounding error}} \approx \frac{\alpha^2}{3\cdot 2^{2b}}.
\end{split}
\end{equation}

By substituting the above equation to the rounding error term, the quantization error is summarized as follows:
\begin{equation}
\begin{gathered}
\text{Quantization Error} = E[(W - Q(W))^2] \\
\approx \overbrace{\cdot \int_{\alpha}^{\infty} f(x)\cdot(x-\alpha)^2dx + \cdot \int_{-\infty}^{-\alpha} f(x)\cdot(x-\alpha)^2dx}^{\text{truncation error}} + \overbrace{\frac{\alpha^2}{3\cdot 2^{2b}}}^{\text{rounding error}},
\end{gathered}
\end{equation}

where $\alpha$ is the truncation boundary that minimizes $||W - Q(W)||_2$. 
In addition, because Gaussian distribution is an even function, two terms of truncation error are identical. Overall, the quantization error is summarized as follows:
\begin{equation}
\begin{split}
&\text{Quantization Error} = E[(W - Q(W))^2] \\
&\approx \overbrace{2 \cdot \int_{\alpha}^{\infty} f(x)\cdot(x-\alpha)^2dx}^{\text{truncation error}} + \overbrace{\frac{\alpha^2}{3\cdot 2^{2b}}}^{\text{rounding error}}.
\end{split}
\end{equation}

In the case of Equation 6, the probability density function of $G(x)$ is the same as $f(x)$, therefore the quantization error in $[-d, d]$ could be achievable following the similar derivation of Equation 10 as given by:
\begin{equation}
\begin{split}
&\text{Quantization Error'} \in (-d, d) \\
&\approx \overbrace{2 \cdot \int_{\alpha'}^{d} f(x)\cdot(x-\alpha')^2dx}^{\text{truncation error}} + \overbrace{\frac{\alpha'^2}{3\cdot 2^{2b}}}^{\text{rounding error}},
\end{split}
\end{equation}
where $\alpha'$ is the truncation boundary that minimizes $||W' - Q(W')||_2$. 
In addition, the error of clamped values is expressed as:
\begin{equation}
\begin{split}
&F(-|d|)(d-\alpha')^2 + (1-F(d))(d-\alpha')^2 \\
&= 2\cdot F(-|d|)\cdot (d-\alpha')^2    
\end{split}
\end{equation}

By combining the above two terms, we can get the overall quantization errors of the clamped weight:
\begin{equation}
\begin{gathered}
\text{Quantization Error}' = E[(W' - Q(W'))^2] \\
\approx \overbrace{2\cdot\Big(F(-|d|)\cdot (d-\alpha')^2 + \int_{\alpha'}^{d} f(x)\cdot(x-\alpha')^2dx\Big)}^{\text{truncation error}} + \overbrace{\frac{\alpha'^2}{3\cdot 2^{2b}}}^{\text{rounding error}}. 
\label{eq:ACIQ_clamped_normal_ERROR}
\end{gathered}
\end{equation}

\subsection{Proof of Error Comparison}
When we compare  Eq. (4) and Eq. (6), Eq. (6) always has a smaller error than Eq. (4), showing that the clamped distribution is more robust than the unbounded distribution. In order to prove the relationship mentioned above, we will first give two lemmas.

\textbf{Lemma 1.} For the arbitrary $A < d$, the quantization error of unbounded distribution with truncation boundary $A$ is always larger than the quantization error of bounded distribution with truncation boundary $A$. 

The difference of error is given as:
\begin{equation}
\begin{split}
&\Delta\text{Error} \\ 
&= 2\cdot \int_{A}^{\infty} f(x)\cdot(x-A)^2dx - 2\cdot F(-|d|)\cdot(d-A)^2 \\
&= 2\cdot \Big(\int_{A}^{\infty} f(x)\cdot(x-A)^2dx - \int_{d}^{\infty} f(x)\cdot(d-A)^2dx\Big) \\
&= 2\cdot \int_{A}^{d} f(x)\cdot(x-A)^2dx \\
& +2\cdot \int_{d}^{\infty} f(x)\big((x-A)^2 - (d-A)^2\big)dx.
\end{split}
\end{equation}
The last two terms are always positive, therefore lemma 1 holds. 

\textbf{Lemma 2.} For the arbitrary $A < d$, the quantization error of bounded distribution with truncation boundary $A$ is always larger than or equal to the quantization error of bounded distribution with truncation boundary $\alpha'$. 

From the definition of $\alpha'$, where $\alpha'$ is the truncation boundary that minimizes $||W' - Q(W')||_2$, lemma 2 is always valid. 

From lemma 1 and lemma 2, for the arbitrary $A < d$, the quantization error of unbounded distribution with truncation boundary $A$ is larger than the quantization error of bounded distribution with truncation boundary $\alpha'$.  Therefore, Eq.(6) is always smaller than to Eq.(4). $\blacksquare$

\bibliographystyle{splncs04}
\bibliography{egbib}